\pdfoutput=1
\documentclass[11pt]{article}

\usepackage[preprint]{acl}

\usepackage{times}
\usepackage{latexsym}
\usepackage{xspace,mfirstuc,tabulary}
\usepackage{float}
\usepackage{times,latexsym}
\usepackage{url}
\usepackage[T1]{fontenc}
\usepackage{tcolorbox}
\usepackage{enumitem}
\usepackage[utf8]{inputenc}
\usepackage{microtype}
\usepackage{caption}
\usepackage{booktabs}
\usepackage{multirow}
\usepackage{framed}
\usepackage{amsmath} 
\usepackage{multicol} % 引入 multicol 宏包
\usepackage{xcolor}   % 用于设置背景颜色
\usepackage{natbib}
\usepackage{tabularx}
\usepackage{bigstrut}
\usepackage{inconsolata}
\usepackage{graphicx}
\definecolor{shadecolor}{gray}{0.9} % 设置背景颜色为浅灰色
 {\endMakeFramed}
\title{GridRoute: A Benchmark for LLM-Based Route Planning with Cardinal Movement in Grid Environments}

\author{
Kechen Li$^1$, Yaotian Tao$^2$, Ximing Wen$^3$, Quanwei Sun$^2$, \protect \\
{\bf Zifei Gong$^4$, Hongli Zhang $^2$, Chang Xu$^2$, Xizhe Zhang$^2$, Tianbo Ji$^{1,}$}\thanks{Corresponding author}
\\$^1$School of Transportation and Civil Engineering, Nantong University, China \\    $^2$HongKong Linchance Intelligent Technology Co., Ltd. (LinChance), China \\$^3$College of Computing and Informatics, Drexel University, USA \\   $^4$University of New South Wales, Australia \\
\texttt{kechen.li@stmail.ntu.edu.cn}, \texttt{\{taoyaotian,quanweisun,HongliZhang,xuchang,zhangxizhe\}@linchance.com} \\
\texttt{xw384@drexel.edu}, \texttt{z3224950@zmail.unsw.edu.au},
\texttt{jitianbo@ntu.edu.cn}
}
\begin{document}
\maketitle

\begin{abstract} 
Recent advancements in Large Language Models (LLMs) have demonstrated their potential in planning and reasoning tasks, offering a flexible alternative to classical pathfinding algorithms. However, most existing studies focus on LLMs’ independent reasoning capabilities and overlook the potential synergy between LLMs and traditional algorithms. To fill this gap, we propose a comprehensive evaluation benchmark GridRoute to assess how LLMs can take advantage of traditional algorithms. We also propose a novel hybrid prompting technique called Algorithm of Planning (AoP), which introduces traditional algorithms' guidance into prompting. Our benchmark evaluates six LLMs ranging from 7B to 72B parameters across various map sizes, assessing their performance in correctness, optimality, and efficiency in grid environments with varying sizes. Our results show that AoP significantly boosts performance across all model sizes, particularly in larger or more complex environments, suggesting a promising approach to addressing path planning challenges.\footnote{Our code is open-sourced at \url{https://github.com/LinChance/GridRoute}.}
\end{abstract}

\section{Introduction}

Pathfinding in grid-based environments has long been dominated by classical algorithms such as A* \cite{Hart1968}, Dijkstra \citep{dijkstra1959note}, and DFS, which rely on explicit search heuristics and guarantee optimality under predefined conditions. However, the recent rise of Large Language Models (LLMs) has introduced an alternative paradigm, everaging their implicit reasoning capabilities and contextual adaptation.

Specialized benchmarks, such as Path Planning from Natural Language (PPNL) \cite{Aglargemodelllmpathplan2024} explored LLMs' ability to perform end-to-end navigation while adhering to movement constraints and avoiding obstacles. However, they only explored the independent exploring ability of LLMs using prompting strategies such as Chain-of-Thought (CoT) \cite{chainofthoughtpromptingelicitsreasoning}, ReAct \cite{yao2023react}.  LLM-A* \cite{meng2025llm}, on the other hand, tried to integrate LLM's ability with guidance from traditional pathfinding algorithms, but they only use LLM in one of the steps, which is to generate intermediate target lists, and still rely on A* to generate the final solution. In this way, there still lacks research exploring how traditional algorithms could help LLM-generated route planning. 

To bridge this gap, we propose a comprehensive benchmark  \textit{GridRoute}  to systematically evaluate the role of traditional algorithmic guidance in enhancing the pathfinding capabilities of LLMs. Our benchmark is designed around three core prompting paradigms: (1) independent prompting, (2) Algorithm of Planning(AoP) prompt, where traditional algorithms such as DFS, A*, and  Dijkstra provide reasoning guidance; and (3) hybrid prompting, which leverages algorithm-generated trajectories or decisions within both direct-answer and reasoning-based prompts. This design facilitates a comprehensive comparison of traditional algorithmic guidance versus independent prompting approaches, and further allows for in-depth analysis of variation within algorithm-guided prompts.

To assess scalability and generalization, we construct grid-based environments of three different sizes with increasing complexity. We evaluate the performance of six LLMs, ranging from 7B to 72B parameters, including models from multiple families, such as GPT-4 \cite{openai2024gpt4technicalreport}, DeepSeek \cite{deepseekai2024deepseekv3technicalreport}, and QWEN \cite{qwen2025qwen25technicalreport}. Model performance is assessed using metrics that capture correctness, optimality, and efficiency—-such as format compliance, path validity, solution quality, error from the optimal path, and runtime. This multifaceted evaluation offers a rigorous analysis of how algorithmic guidance affects LLM pathfinding, both in terms of accuracy and efficiency.

Through this benchmark, we aim to answer several key questions: To what extent can LLMs benefit from classical search heuristics? How do the different shot strategies affect the traditional algorithm's guidance for LLM? Do algorithm-guided prompt performance remain indispensable regardless of model scale and map size? Our findings suggest that algorithm-guided prompts achieve better results than vanilla prompts and comparable performance compared with the CoT prompt. While larger models tend to perform better under independent prompting, algorithm-guided strategies significantly boost performance across all model sizes, particularly in larger or more complex environments. Moreover, integrating few-shot learning could further improve the performance of algorithm-guided prompt. These results highlight the complementary strengths of LLMs and classical algorithms and point toward promising directions for hybrid neuro-symbolic planning systems.

In conclusion, this study makes three major contributions:
\begin{itemize}
    \item We propose GridRoute, a comprehensive benchmark designed to evaluate LLMs' route planning abilities in grid-based environments, comparing independent prompting and algorithm-guided strategies.
    \item We introduce AoP prompting, a novel framework that guides LLMs using classical pathfinding algorithms (e.g., A*, Dijkstra, DFS) by embedding their reasoning principles directly into the prompt.
    \item We perform large-scale experiments across multiple LLMs and environment sizes, showing that AoP consistently improves planning performance, especially in complex settings, and offers complementary benefits to traditional CoT prompting.
\end{itemize}

\section{Related Work}
\subsection{Grid-based route planning}
Grid-based route planning is central to robotics and embedded systems, especially in warehouse automation, where robots navigate structured indoor environments such as aisles and shelves. These environments are typically modeled as 2D grids with discrete free or occupied cells, offering memory and computational efficiency suitable for embedded platforms. Algorithms such as A* \citep{Hart1968}, Dijkstra’s \cite{dijkstra1959note}, and variants like Jump Point Search \citep{10.5555/3038794.3038810} are widely used for computing collision-free, cost-effective paths. Many systems adopt 4-directional movement (up, down, left, right) to reflect real-world layouts and simplify pathfinding. This constraint, along with lightweight planning algorithms, enables real-time decision-making under tight power and timing constraints, making grid-based navigation viable on low-power processors.
\begin{figure*}[!h]
    \centering
    \includegraphics[width=\linewidth]{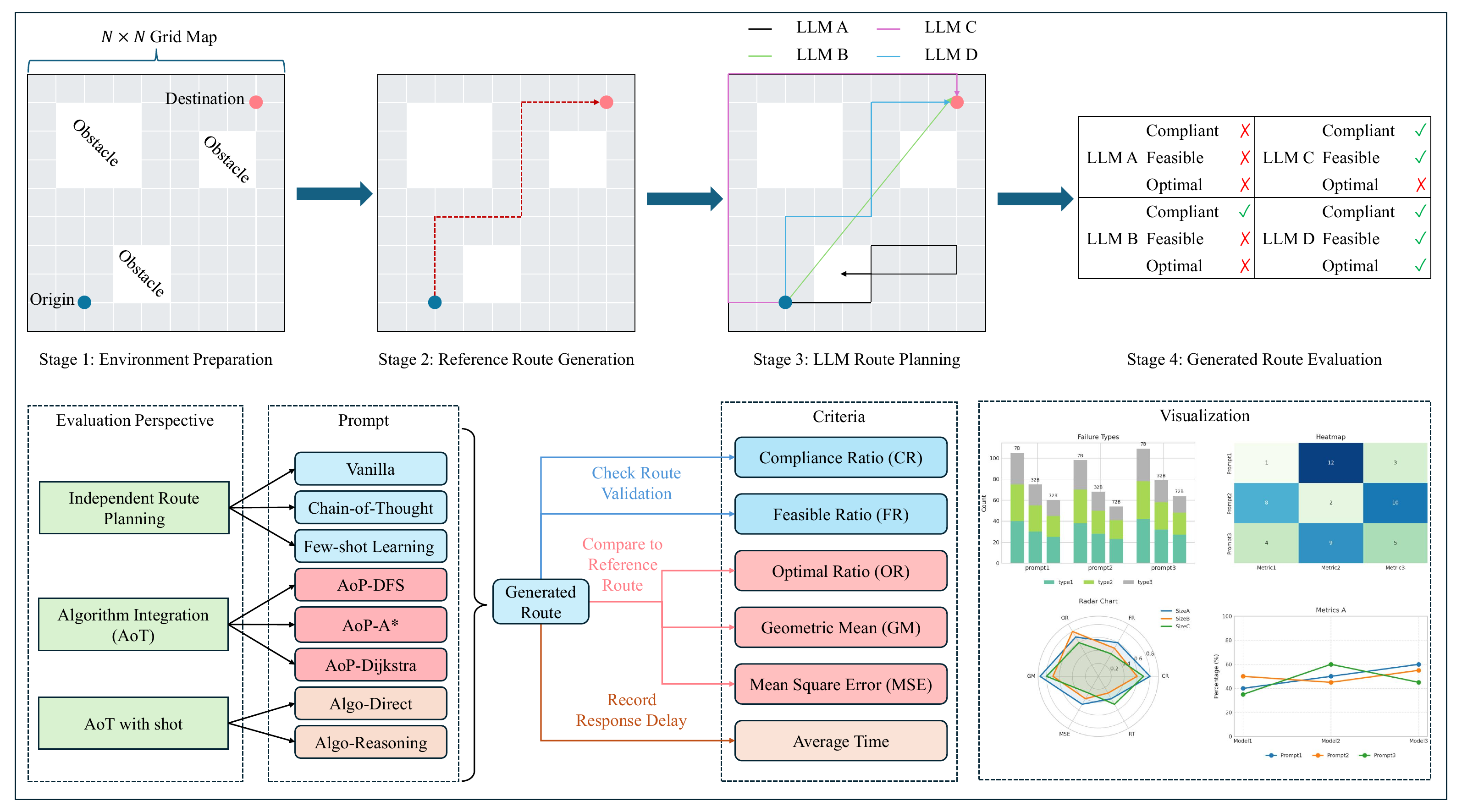}
    \caption{Struture of GridRoute}
    \label{fig:struture}
\end{figure*}
\subsection{Route planning with LLM}
Recent advances in large language models (LLMs) have showcased their emerging capabilities in reasoning, planning, and problem-solving. Unlike classical algorithms, LLMs leverage implicit knowledge and contextual understanding, allowing them to generate flexible solutions in unfamiliar scenarios \citep{wang2023documentlevelmachinetranslationlarge, li2025sos1o1r1likereasoning, wen2025language}. Early studies have demonstrated that LLMs can produce plausible paths or high-level plans in grid-based environments when provided with textual descriptions of the environment, along with specified start and goal positions \citep{valmeekam2023planningabilitieslargelanguage, latif20243p}. However, these outputs are often inconsistent or suboptimal, particularly as the complexity of the environment increases.

To address these limitations, recent research has introduced structured reasoning techniques to improve LLM-based planning. For example, the PPNL benchmark investigates the use of Chain-of-Thought (CoT) and ReAct prompting to encourage step-by-step reasoning in route planning tasks. While these methods improve performance over naive prompting, they still do not fully leverage the systematic search capabilities of classical algorithms. In parallel, an emerging line of work explores hybrid approaches that integrate LLMs with traditional pathfinding algorithms such as A* and Dijkstra \citep{liu2023llm+}. In these settings, LLMs may generate sub-goals for symbolic planners to solve or invoke classical algorithms as external tools within a CoT-style reasoning framework.

Building on this line of work, our study systematically investigates how classical algorithms can be used to guide LLM-based route planners. We focus on grid-based environments, where structured search procedures are well-defined, computationally efficient, and particularly suited for real-time or embedded applications.
\section{GridRoute Benchmark}

% This section provides a detailed description of the GridRoute benchmark, including th and the DijPrompt approach.

This section introduces the design of the GridRoute benchmark, including data generation, evaluation metrics, and error analysis. We first describe how the task environments are constructed and presented to LLMs, followed by the metrics used to assess performance. Finally, we analyze common failure modes to better understand the limitations of LLMs in route planning tasks.

\subsection{Data Generation}

GridRoute is proposed to evaluate the performance of LLMs on the task of automated route planning, and its overall structure is illustrated in Figure \ref{fig:struture}. In GridRoute, LLMs are required to discover a valid path from the origin $O$ to the destination $D$ in a randomly generated $N \times N$ grid environment, while avoiding a set of $n$ non-overlapping square obstacles of size $s \times s$.  The movement of the agent is constrained to four cardinal directions: up, down, left, and right.

 \textbf{Environment Setting} To give a comprehensive evaluation of different prompting strategies, we employ three sets of map configurations in our experiments with spatial layouts of varying complexity: $\{N=10, n=2, s=3\}$, $\{N=20, n=3, s=4\}$, and $\{N=30, n=4, s=5\}$. For each generated environment, five pairs of origin and destination points are randomly sampled. To ensure task difficulty and meaningful planning, each pair must satisfy a minimum Euclidean distance constraint of at least 30\% of the grid’s diagonal length. In addition, to ensure path feasibility, we verify connectivity using a four-directional Breadth-First Search (BFS) algorithm.      

\textbf{Task Description} All environments and tasks are presented to LLMs in natural language. To this end, we design standardized templates to verbalize structured task information, including grid size, obstacle locations, and origin/destination coordinates, into consistent and interpretable natural language prompts. For evaluation, we employ a constrained diagonal Dijkstra algorithm to compute the shortest paths for all task instances, yielding both the reference trajectory and the corresponding optimal path length.

Although this study focuses on the three predefined configurations, the GridRoute framework is inherently extensible and supports flexible adjustments to map size, obstacle density, and obstacle shapes. This design encourages future exploration of LLMs’ generalization capabilities in larger and more complex planning scenarios.

\subsection{Evaluation Metrics}
We evaluate the benchmark from six perspectives to assess the accuracy, quality, and efficiency of LLM-generated routes, as detailed below:

\textbf{Compliance Ratio (CR):} It indicates whether an LLM can produce a specific formatted output as instructed, namely a route in this case. A continuous route which starts from the origin and ends at the destination can be deemed ``compliant''. It can be simply computed as $\text{CR} = \frac{\text{N}_{comp}}{\text{N}_{case}} \times 100\%$ where $\text{N}_{comp}$ and $\text{N}_{case}$ are the numbers of compliant routes and test cases, respectively.
 
 \textbf{Feasibility Ratio (FR):} It shows whether an LLM can generate a feasible route, where a ``feasible'' route is ``compliant'' and avoid any obstacle in it. FR is computed as $\text{FR} = \frac{\text{N}_{feas}}{\text{N}_{case}} \times 100\%$.

\textbf{Optimal Ratio (OR):} It is used to assess whether a feasible LLM-generated route is optimal compared to the Dijkstra ground truth according to the length. OR is computed as $\text{OR} = \frac{\text{N}_{optm}}{\text{N}_{case}} \times 100\%$.

\textbf{Geometric Mean (GM):} It measures the similarity between LLM-generated routes and reference optimal routes. It is computed as $\text{GM} = \exp \left( \frac{1}{n} \sum\limits_{i=1}^{n} \log(\frac{len(\text{R}^{\text{M}}_{i})}{len(\text{R}^{\text{R}}_{i})} ) \right)$ where $len$ returns the length of a given route, $\text{R}^{\text{M}}$ and $\text{R}^{\text{R}}$ are the LLM-generated route and the Dijkstra ground-truth route, respectively. Higher LLM performance is associated with GM approaching 1.

\textbf{Mean Square Error (MSE):} It measures the difference between candidate routes and optimal routes, computed as $\text{MSE} = \frac{1}{n} \sum\limits_{i=1}^{n} \left(len(\text{R}^{\text{M}}_{i}) - len(\text{R}^{\text{R}}_{i})\right)^2$. MSE closer to 0 indicates a higher performance.

\textbf{Run Time (RT(s)):}It records the average time for LLMs to generate paths of different sizes.

\subsection{Route Failure Types}
As part of the GridRoute benchmark, we further categorize the generated failure cases into five representative error types, including:

 \textbf{Invalid Step Distance:} Indicates the presence of non-unit-length steps (e.g., diagonal moves) in the path, violating grid-based movement constraints.
 
 \textbf{Path Through Obstacle:}  Occurs when the generated path intersects predefined obstacle regions, failing to satisfy collision-free requirements.  
 
 \textbf{Out of Bounds:}  Reflects paths exceeding map boundaries, revealing the LLMs' lack of spatial awareness regarding environmental constraints.
 
 \textbf{Empty Path:} Denotes failure cases where the LLM produces no path output. 
 
 \textbf{Start/End Mismatch:} Identifies paths that either deviate from the specified origin or fail to reach the designated destination. 

\section{Experiments}

\subsection{Experiments Setup}
In our experiments, we conducted a comprehensive evaluation involving the Qwen2.5 series, LLaMA3.1-70B \citep{grattafiori2024llama3herdmodels}, DeepSeek-V3, and GPT-4 Turbo. The Qwen2.5 series includes models of various parameter scales, specifically 7B, 32B, and 72B while DeepSeek-V3 has 671 billion parameters.

\subsection{Prompt Design}
To evaluate how the guidance of traditional algorithms influences the ability to plan the route of LLMs, we experiment with mainly three categories of prompting strategies, with examples of each prompt shown in the appendix \ref{appendix:01}.
\subsubsection{Independent Route Planning Prompt}
In this part, we use three types of prompts to evaluate the independent planning capabilities of LLMs: vanilla instruction-based prompting, CoT prompting, and Few-shot Learning prompting. 

\
 \textbf{Vanilla:} it is the basic prompt which provides straightforward instructions for LLMs to complete the path planning task, by which LLMs are simply asked to generate a continuous, obstacle-free route given the map size and the position coordinates of the origin and destination. 
 
 \textbf{CoT:} This prompt guides the model through a structured planning procedure. It instructs the LLM to first validate that the start and end points are not blocked, identify valid moves, and iteratively construct the path. This design encourages the model to explicitly reason about each step of the planning process.

\textbf{FewShot-Base:} It presents two examples with complete input-output pairs, including obstacle descriptions and optimal paths, allowing the model to learn the structure of the task by analogy.

\subsubsection{Algorithms of Planning (AoP) Prompt}
To evaluate the capability of LLMs in applying classical algorithms, we designed the Algorithms of Planning (AoP) prompting framework. This framework guides LLMs to mimic classical path planning algorithms by prompting them to apply algorithmic reasoning, analyze the task environment, constrain movement directions, explain algorithmic principles, and adhere to a clear output format.

\textbf{AoP-DFS:} This prompt is designed to guide the model in following the Depth-First Search (DFS).Instead of outlining the algorithm formally, the prompt incorporates its core mechanisms—ordered expansion, backtracking, and visit marking—into the structure of the instruction itself. By structuring the order of exploration and discouraging revisits,it leads the model to follow a behavior pattern similar to DFS.

 \textbf{AoP-A*:} This prompt adapts the logic of the A* algorithm to guide the model in path planning. It introduces key elements of A*, such as cost estimation, heuristic evaluation, and goal-directed prioritization. The prompt defines valid moves, explains how to measure distance using the Manhattan heuristic, and describes how to compare path options. The model is prompted to evaluate each possible move based on its distance to the goal and the cost incurred so far, simulating how A* selects efficient paths.
 
 \textbf{AoP-Dijkstra:} It provides a structured framework by embedding Dijkstra's core logic -- greedy distance minimization, neighbor evaluation, and path optimality. In detail, it enforces a multi-step workflow: initializing Dijkstra's algorithm (infinite distances except the starting cell), iteratively updating route costs via priority queue simulation, and reconstructing the shortest route through parent-node backtracking. 

\subsubsection{AoP with Example (Algo-Shot) Prompt} 
To evaluate the impact of integrating LLMs with traditional algorithms under different shot strategies, we design two distinct shot prompt configurations as follows:

 \textbf{Algo-Direct:} It integrates the AoP-Dijkstra prompt with the same example used in FewShot-Base - providing two cases in the $10 \times 10$ grid, with each case including only the start point, end point, obstacles, and the optimal path computed using the Dijkstra algorithm.
  
\textbf{Algo-Reasoning:} Different from Algo-Direct, Algo-Reasoning combines AoP-Dijkstra with example that provides a detailed demonstration of the path planning reasoning steps from the starting point to the destination. To align with the reasoning process in this example, we configured the internal steps of AoP-Dijkstra in greater detail, clearly specifying how various situations encountered during path planning are handled. 
\begin{figure}[ht!]
    \centering
    \includegraphics[width=\linewidth]{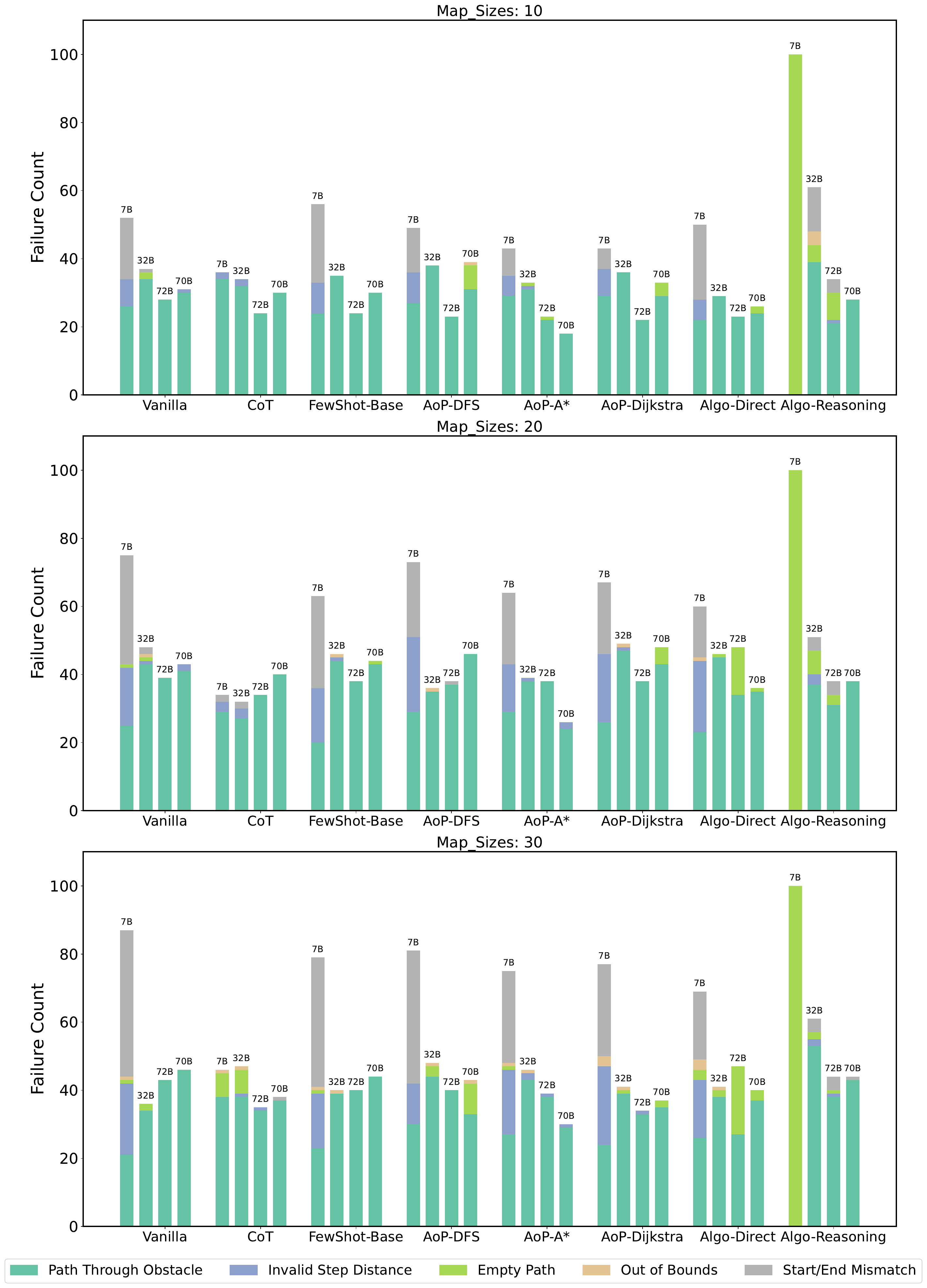}
    \caption{Error distribution of LLMs across three map sizes for route planning. Each subplot represents a map size, with prompts on the x-axis and stacked error counts on the y-axis. Model sizes (7B, 32B, 72B for Qwen2.5; 70B for LLaMA3.1) are labeled above each bar.}
    \label{fig:failure_reasons_by_matrix_size.png}
\end{figure}

\section{Experimental Analysis}

In this section, building upon the four core research questions (RQ) introduced in the Introduction, we conduct the following analyses to provide deeper insights into how various prompting strategies interact with traditional algorithmic guidance under varying conditions.

\begin{figure*}[ht!]
    \centering
    \includegraphics[width=\linewidth]{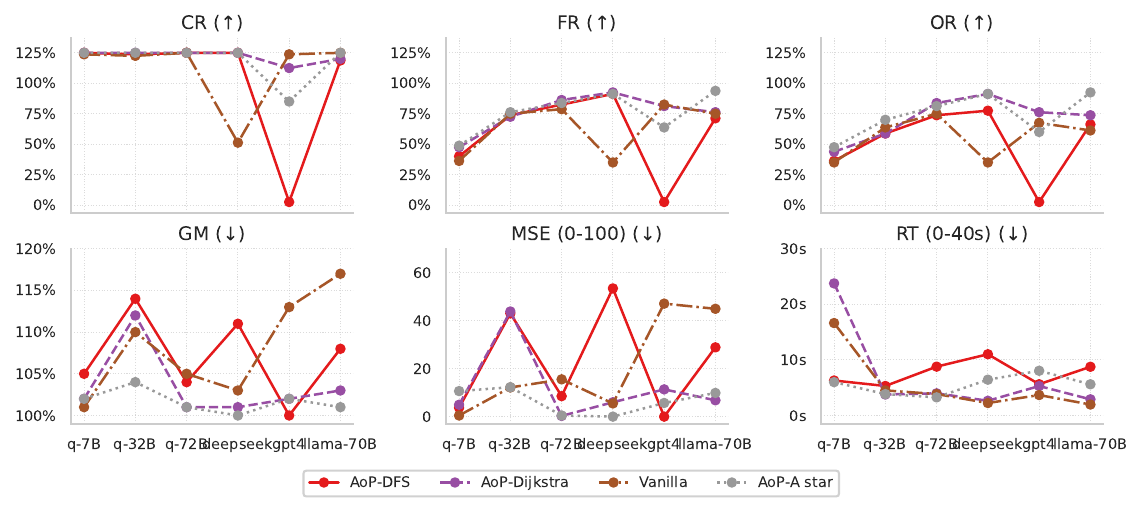}
    \caption{Average evaluation results across three map sizes for each metric using Vanilla and AoP prompts (q-7B: Qwen2.5-7B; q-32B: Qwen2.5-32B; q-72B: Qwen2.5-72B)}
    \label{fig:performance_metrics_vanilla}
\end{figure*}

\begin{figure*}[ht!]
    \centering
    \includegraphics[width=\linewidth]{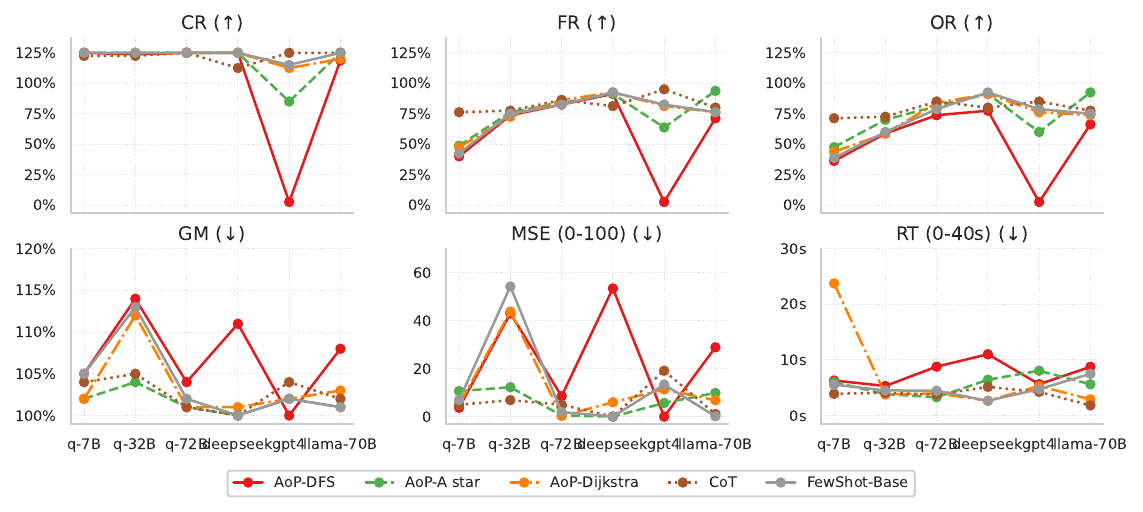}
    \caption{Average evaluation results across three map sizes for each metric using CoT, FewShot-Base, and AoP prompts.}
    \label{fig:performance_metrics_cot}
\end{figure*}

% \begin{figure}[h!]
%     \centering
%     \includegraphics[width=\linewidth]{./photo/aot_heatmap.pdf}
%     \caption{AoT prompt win rates across each evaluation metric, with darker colors indicating higher values.}
%     \label{fig:aot_heatmap}
% \end{figure}

\begin{figure*}[ht!]
    \centering
    \includegraphics[width=\linewidth]{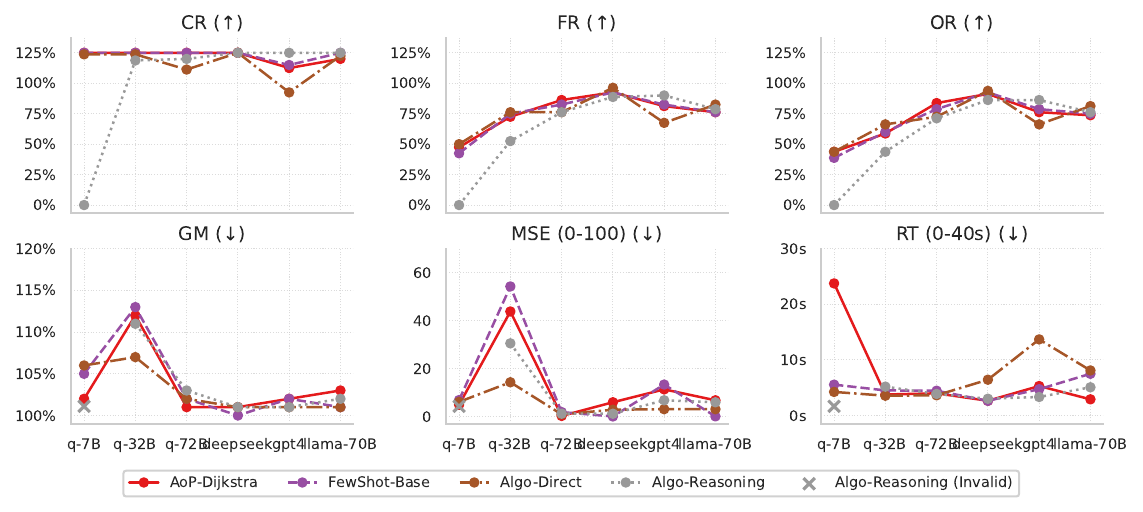}
    \caption{Average evaluation results across three map sizes for each metric using FewShot-Base and Algo-Shot prompts.}
    \label{fig:performance_metrics_shot}
\end{figure*}

\begin{figure*}[h!]
    \centering
    \includegraphics[width=\linewidth]{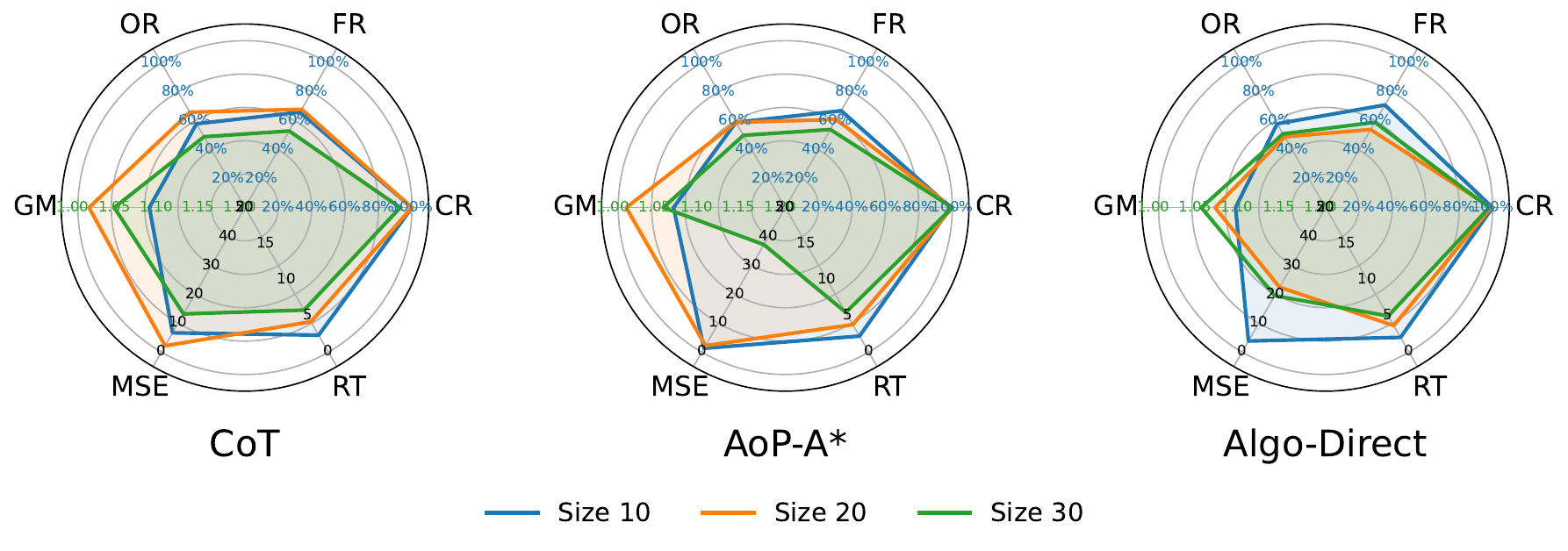}
    \caption{Performance evaluation of Qwen2.5-32B using three prompt strategies (CoT, AoP-A*, Algorithm-Direct) across multiple map sizes and metrics.}
    \label{fig:qwen2.5-32B_comparision}
\end{figure*}

\textbf{RQ1: How does the Algorithms of Planning (AoP) prompt perform compared to the Independent Route Planning prompt?}

% AoT和Vanilla对比
In Figure \ref{fig:performance_metrics_vanilla}, we find that AoP prompting methods significantly outperform the Vanilla prompt across all metrics except AoP-DFS, especially in more complex environments. Moreover, the integration of LLMs with optimal pathfinding algorithms (e.g., Dijkstra or A*) achieves better planning performance compared to their integration with uninformed search methods like DFS. These methods show clear advantages in terms of FR and OR, indicating stronger path planning capabilities. However, a notable trade-off exists for AoP-DFS compared to Vanilla: despite the improvements in FR and OR, GM and MSE can sometimes be higher.

%AoT和CoT/Few-shot Learning对比
Compared to the more advanced independent prompting strategy CoT, AoP achieves comparable or even superior performance, particularly on Qwen2.5-7B, as illustrated in Figure~\ref{fig:performance_metrics_cot}. 

% Figure~\ref{fig:aot_heatmap} further highlights that AoT surpasses CoT in terms of geometric mean (GM), reflecting higher overall route quality. 

However, CoT exhibits more consistent performance across all models, with fewer failure cases as shown in Figure~\ref{fig:failure_reasons_by_matrix_size.png}. In contrast, within the AoP framework, only AoP-Dijkstra consistently maintains robust results. These findings suggest that CoT’s step-by-step reasoning offers greater reliability across model sizes, while AoP’s effectiveness depends more on the model’s ability to accurately internalize and apply classical algorithmic logic.

FewShot-Base, despite its simplicity compared to advanced prompting strategies such as CoT and AoP, achieves performance comparable to AoP-Dijkstra by relying on minimal task demonstrations involving only obstacles, start/end points, and optimal paths. This highlights the efficacy of lightweight imitation-based learning and illustrates the strong transfer and generalization capabilities of LLMs in path planning tasks.

Overall, CoT demonstrates the lowest total error count, while AoP prompts shows comparable performance overall. However, AoP exhibits significant start/end mismatch issues in smaller models (e.g., Qwen2.5-7B).

\begin{tcolorbox}[title=Remark 1]
 AoP significantly improves LLM route planning performance compared with Vanilla. AoP-Dijkstra and CoT maintain robust performance across different models and map sizes without introducing significant inference latency, making them the most generally stable prompting methods.
\end{tcolorbox}

\textbf{RQ2: How few shots influence LLM's planning ability when guided by traditional algorithms?}

Our experiments comparing AoP-Dijkstra with different shot strategies (Figure \ref{fig:performance_metrics_shot}) reveal that the Algo-Direct prompt generally enhances AoP-Dijkstra performance in most metrics, while the Algo-Reasoning setting improves its performance specifically in GM and MSE at the cost of reduced FR and OR. However, due to the limited capabilities of Qwen2.5-7B, Algo-Reasoning could not produce valid outputs on this smaller model. When using the same example, Algo-Direct prompting outperforms  FewShot-Base by using the AoP-Dijkstra prompt while preserving example demonstration patterns.

\begin{tcolorbox}[title=Remark 2]
Direct examples without reasoning with AoP tend to yield more consistent performance, suggesting that example complexity may influence the model's ability to generalize under structured prompting frameworks.
\end{tcolorbox}

\textbf{RQ3: How does the size of the model influence the traditional algorithm-guided prompt?}

Figure \ref{fig:failure_reasons_by_matrix_size.png} demonstrates that within the Qwen2.5 model series, scaling from 7B to 32B leads to substantial improvements in FR and RT. The 72B model further enhances FR while significantly boosting OR and reducing GM and MSE. This suggests that the FR gains observed in Qwen2.5-32B come at the cost of higher GM and MSE. Notably, Qwen2.5-72B achieves performance comparable to models such as GPT-4 Turbo(1 trillion) and DeepSeek-V3(671B), suggesting that \textbf{simply increasing model size in GridRoute tasks eventually encounters a scaling law bottleneck}. Additionally, as model size increases, errors such as illegal moves and start-end mismatches become significantly less frequent and are increasingly concentrated in path-through-obstacle cases. In most prompting strategies, inference speed tends to improve with larger model sizes, especially under AoP-Dijkstra and vanilla prompts. However, AoP-DFS shows less consistent improvement and even experiences relatively increased latency from 7B to 32B, indicating that response time is not strictly correlated with model size.

\begin{tcolorbox}[title=Remark 3]
Scaling up model size in GridRoute tasks generally improves accuracy and reduces errors, but performance gains plateau beyond a certain size, indicating diminishing returns. Conversely, inference time tends to decrease as model size increases.
\end{tcolorbox}

\textbf{RQ4: How does map size influence traditional algorithm-guided prompt?}

As shown in Figure \ref{fig:qwen2.5-32B_comparision}, the scaling of map size is a critical factor affecting the path planning performance across all evaluation metrics. As the map size increases, both the FR and OR exhibit a systematic decline, while the RT correspondingly increases. This scaling effect holds consistently regardless of model size, although the influence of model size diminishes as the map becomes larger—for instance, on a 30×30 grid, the performance of Qwen2.5-32B approaches that of Qwen2.5-72B. Notably, the nature of errors evolves with increasing map complexity, transitioning from primarily " Path though obstacles" to more intricate types of errors. While model size and architectural differences do influence performance, their impact is secondary to the fundamental constraints imposed by map dimensionality. In particular, the CoT prompting strategy exhibits more generalized adaptability across varying map sizes compared to the AoP approach. This suggests that with appropriate prompting strategies, LLMs can achieve path planning performance comparable to or even competitive with AoP methods.
\begin{tcolorbox}[title=Remark 4]
 Map size is the dominant factor impacting path planning performance, with larger maps diminishing the advantages of model scaling; compared to map size, model size has a smaller influence to performance.
% , while CoT prompting offers greater adaptability across varying complexities.
\end{tcolorbox}

\section{Conclusion}
We introduced GridRoute, a benchmark for systematically evaluating LLM-based route planning with guidance from classical algorithms. Our proposed AoP prompting framework enables LLMs to incorporate structured reasoning inspired by A*, Dijkstra, and DFS, significantly improving performance compared with Vallina, especially in complex environments. AoP performs comparably to or better than CoT, while example-based AoP variants further enhance generalization. Although CoT offers more stable performance across model sizes, our findings highlight the complementary strengths of symbolic algorithms and LLMs, suggesting promising directions for hybrid neuro-symbolic planning systems.
\section{Limitations}
 Our current benchmark shows that the performance of 32B and 70B models tends to converge as map dimensions increase. However, the performance limits of LLMs on ultra-large-scale maps remain unclear and require extended experiments to assess the trade-off between computational efficiency and planning accuracy. Notably, while increasing model size and refining prompting strategies significantly reduce most types of error, critical issues such as path traversal through obstacles persist. These could potentially be mitigated through mechanisms like obstacle collision backtracking or targeted fine-tuning. Furthermore, existing studies focus solely on single-goal planning in open maps. The GridRoute framework, however, can be naturally extended to multi-goal cooperative planning scenarios—such as logistics delivery or multi-task inspection—marking a promising direction for hybrid decision-making that integrates traditional algorithms with LLMs.

% \section*{Acknowledgments}

% This document has been adapted
% by Steven Bethard, Ryan Cotterell and Rui Yan
% from the instructions for earlier ACL and NAACL proceedings, including those for
% ACL 2019 by Douwe Kiela and Ivan Vuli\'{c},
% NAACL 2019 by Stephanie Lukin and Alla Roskovskaya,
% ACL 2018 by Shay Cohen, Kevin Gimpel, and Wei Lu,
% NAACL 2018 by Margaret Mitchell and Stephanie Lukin,
% Bib\TeX{} suggestions for (NA)ACL 2017/2018 from Jason Eisner,
% ACL 2017 by Dan Gildea and Min-Yen Kan,
% NAACL 2017 by Margaret Mitchell,
% ACL 2012 by Maggie Li and Michael White,
% ACL 2010 by Jing-Shin Chang and Philipp Koehn,
% ACL 2008 by Johanna D. Moore, Simone Teufel, James Allan, and Sadaoki Furui,
% ACL 2005 by Hwee Tou Ng and Kemal Oflazer,
% ACL 2002 by Eugene Charniak and Dekang Lin,
% and earlier ACL and EACL formats written by several people, including
% John Chen, Henry S. Thompson and Donald Walker.
% Additional elements were taken from the formatting instructions of the \emph{International Joint Conference on Artificial Intelligence} and the \emph{Conference on Computer Vision and Pattern Recognition}.

% Bibliography entries for the entire Anthology, followed by custom entries
%\bibliography{anthology,custom}
% Custom bibliography entries only
\bibliography{main}

\begin{thebibliography}{17}
\providecommand{\natexlab}[1]{#1}

\bibitem[{Aghzal et~al.(2023)Aghzal, Plaku, and Yao}]{Aglargemodelllmpathplan2024}
Mohamed Aghzal, Erion Plaku, and Ziyu Yao. 2023.
\newblock Can large language models be good path planners? a benchmark and investigation on spatial-temporal reasoning.
\newblock \emph{arXiv preprint arXiv:2310.03249}.

\bibitem[{DeepSeek-AI et~al.(2024)DeepSeek-AI, Liu, Feng, Xue, Wang, Wu, Lu, Zhao, Deng, Zhang, Ruan, Dai, Guo, Yang, Chen, Ji, Li, Lin, Dai, Luo, Hao, Chen, Li, Zhang, Bao, Xu, Wang, Zhang, Ding, Xin, Gao, Li, Qu, Cai, Liang, Guo, Ni, Li, Wang, Chen, Chen, Yuan, Qiu, Li, Song, Dong, Hu, Gao, Guan, Huang, Yu, Wang, Zhang, Xu, Xia, Zhao, Wang, Zhang, Li, Wang, Zhang, Zhang, Tang, Li, Tian, Huang, Wang, Zhang, Wang, Zhu, Chen, Du, Chen, Jin, Ge, Zhang, Pan, Wang, Xu, Zhang, Chen, Li, Lu, Zhou, Chen, Wu, Ye, Ye, Ma, Wang, Zhou, Yu, Zhou, Pan, Wang, Yun, Pei, Sun, Xiao, Zeng, Zhao, An, Liu, Liang, Gao, Yu, Zhang, Li, Jin, Wang, Bi, Liu, Wang, Shen, Chen, Zhang, Chen, Nie, Sun, Wang, Cheng, Liu, Xie, Liu, Yu, Song, Shan, Zhou, Yang, Li, Su, Lin, Li, Wang, Wei, Zhu, Zhang, Xu, Xu, Huang, Li, Zhao, Sun, Li, Wang, Yu, Zheng, Zhang, Shi, Xiong, He, Tang, Piao, Wang, Tan, Ma, Liu, Guo, Wu, Ou, Zhu, Wang, Gong, Zou, He, Zha, Xiong, Ma, Yan, Luo, You, Liu, Zhou, Wu, Ren, Ren, Sha, Fu, Xu, Huang, Zhang, Xie, Zhang, Hao,
  Gou, Ma, Yan, Shao, Xu, Wu, Zhang, Li, Gu, Zhu, Liu, Li, Xie, Song, Gao, and Pan}]{deepseekai2024deepseekv3technicalreport}
DeepSeek-AI, Aixin Liu, Bei Feng, Bing Xue, Bingxuan Wang, Bochao Wu, Chengda Lu, Chenggang Zhao, Chengqi Deng, Chenyu Zhang, Chong Ruan, Damai Dai, Daya Guo, Dejian Yang, Deli Chen, Dongjie Ji, Erhang Li, Fangyun Lin, Fucong Dai, and 181 others. 2024.
\newblock \href {https://arxiv.org/abs/2412.19437} {Deepseek-v3 technical report}.
\newblock \emph{Preprint}, arXiv:2412.19437.

\bibitem[{Dijkstra(1959)}]{dijkstra1959note}
Edsger~W Dijkstra. 1959.
\newblock A note on two problems in connexion with graphs.
\newblock \emph{Numerische mathematik}, 1(1):269--271.

\bibitem[{Grattafiori et~al.(2024)Grattafiori, Dubey, Jauhri, Pandey, Kadian, Al-Dahle, Letman, Mathur, Schelten, Vaughan, Yang, Fan, Goyal, Hartshorn, Yang, Mitra, Sravankumar, Korenev, Hinsvark, Rao, Zhang, Rodriguez, Gregerson, Spataru, Roziere, Biron, Tang, Chern, Caucheteux, Nayak, Bi, Marra, McConnell, Keller, Touret, Wu, Wong, Ferrer, Nikolaidis, Allonsius, Song, Pintz, Livshits, Wyatt, Esiobu, Choudhary, Mahajan, Garcia-Olano, Perino, Hupkes, Lakomkin, AlBadawy, Lobanova, Dinan, Smith, Radenovic, Guzmán, Zhang, Synnaeve, Lee, Anderson, Thattai, Nail, Mialon, Pang, Cucurell, Nguyen, Korevaar, Xu, Touvron, Zarov, Ibarra, Kloumann, Misra, Evtimov, Zhang, Copet, Lee, Geffert, Vranes, Park, Mahadeokar, Shah, van~der Linde, Billock, Hong, Lee, Fu, Chi, Huang, Liu, Wang, Yu, Bitton, Spisak, Park, Rocca, Johnstun, Saxe, Jia, Alwala, Prasad, Upasani, Plawiak, Li, Heafield, Stone, El-Arini, Iyer, Malik, Chiu, Bhalla, Lakhotia, Rantala-Yeary, van~der Maaten, Chen, Tan, Jenkins, Martin, Madaan, Malo, Blecher,
  Landzaat, de~Oliveira, Muzzi, Pasupuleti, Singh, Paluri, Kardas, Tsimpoukelli, Oldham, Rita, Pavlova, Kambadur, Lewis, Si, Singh, Hassan, Goyal, Torabi, Bashlykov, Bogoychev, Chatterji, Zhang, Duchenne, Çelebi, Alrassy, Zhang, Li, Vasic, Weng, Bhargava, Dubal, Krishnan, Koura, Xu, He, Dong, Srinivasan, Ganapathy, Calderer, Cabral, Stojnic, Raileanu, Maheswari, Girdhar, Patel, Sauvestre, Polidoro, Sumbaly, Taylor, Silva, Hou, Wang, Hosseini, Chennabasappa, Singh, Bell, Kim, Edunov, Nie, Narang, Raparthy, Shen, Wan, Bhosale, Zhang, Vandenhende, Batra, Whitman, Sootla, Collot, Gururangan, Borodinsky, Herman, Fowler, Sheasha, Georgiou, Scialom, Speckbacher, Mihaylov, Xiao, Karn, Goswami, Gupta, Ramanathan, Kerkez, Gonguet, Do, Vogeti, Albiero, Petrovic, Chu, Xiong, Fu, Meers, Martinet, Wang, Wang, Tan, Xia, Xie, Jia, Wang, Goldschlag, Gaur, Babaei, Wen, Song, Zhang, Li, Mao, Coudert, Yan, Chen, Papakipos, Singh, Srivastava, Jain, Kelsey, Shajnfeld, Gangidi, Victoria, Goldstand, Menon, Sharma, Boesenberg,
  Baevski, Feinstein, Kallet, Sangani, Teo, Yunus, Lupu, Alvarado, Caples, Gu, Ho, Poulton, Ryan, Ramchandani, Dong, Franco, Goyal, Saraf, Chowdhury, Gabriel, Bharambe, Eisenman, Yazdan, James, Maurer, Leonhardi, Huang, Loyd, Paola, Paranjape, Liu, Wu, Ni, Hancock, Wasti, Spence, Stojkovic, Gamido, Montalvo, Parker, Burton, Mejia, Liu, Wang, Kim, Zhou, Hu, Chu, Cai, Tindal, Feichtenhofer, Gao, Civin, Beaty, Kreymer, Li, Adkins, Xu, Testuggine, David, Parikh, Liskovich, Foss, Wang, Le, Holland, Dowling, Jamil, Montgomery, Presani, Hahn, Wood, Le, Brinkman, Arcaute, Dunbar, Smothers, Sun, Kreuk, Tian, Kokkinos, Ozgenel, Caggioni, Kanayet, Seide, Florez, Schwarz, Badeer, Swee, Halpern, Herman, Sizov, Guangyi, Zhang, Lakshminarayanan, Inan, Shojanazeri, Zou, Wang, Zha, Habeeb, Rudolph, Suk, Aspegren, Goldman, Zhan, Damlaj, Molybog, Tufanov, Leontiadis, Veliche, Gat, Weissman, Geboski, Kohli, Lam, Asher, Gaya, Marcus, Tang, Chan, Zhen, Reizenstein, Teboul, Zhong, Jin, Yang, Cummings, Carvill, Shepard, McPhie,
  Torres, Ginsburg, Wang, Wu, U, Saxena, Khandelwal, Zand, Matosich, Veeraraghavan, Michelena, Li, Jagadeesh, Huang, Chawla, Huang, Chen, Garg, A, Silva, Bell, Zhang, Guo, Yu, Moshkovich, Wehrstedt, Khabsa, Avalani, Bhatt, Mankus, Hasson, Lennie, Reso, Groshev, Naumov, Lathi, Keneally, Liu, Seltzer, Valko, Restrepo, Patel, Vyatskov, Samvelyan, Clark, Macey, Wang, Hermoso, Metanat, Rastegari, Bansal, Santhanam, Parks, White, Bawa, Singhal, Egebo, Usunier, Mehta, Laptev, Dong, Cheng, Chernoguz, Hart, Salpekar, Kalinli, Kent, Parekh, Saab, Balaji, Rittner, Bontrager, Roux, Dollar, Zvyagina, Ratanchandani, Yuvraj, Liang, Alao, Rodriguez, Ayub, Murthy, Nayani, Mitra, Parthasarathy, Li, Hogan, Battey, Wang, Howes, Rinott, Mehta, Siby, Bondu, Datta, Chugh, Hunt, Dhillon, Sidorov, Pan, Mahajan, Verma, Yamamoto, Ramaswamy, Lindsay, Lindsay, Feng, Lin, Zha, Patil, Shankar, Zhang, Zhang, Wang, Agarwal, Sajuyigbe, Chintala, Max, Chen, Kehoe, Satterfield, Govindaprasad, Gupta, Deng, Cho, Virk, Subramanian, Choudhury,
  Goldman, Remez, Glaser, Best, Koehler, Robinson, Li, Zhang, Matthews, Chou, Shaked, Vontimitta, Ajayi, Montanez, Mohan, Kumar, Mangla, Ionescu, Poenaru, Mihailescu, Ivanov, Li, Wang, Jiang, Bouaziz, Constable, Tang, Wu, Wang, Wu, Gao, Kleinman, Chen, Hu, Jia, Qi, Li, Zhang, Zhang, Adi, Nam, Yu, Wang, Zhao, Hao, Qian, Li, He, Rait, DeVito, Rosnbrick, Wen, Yang, Zhao, and Ma}]{grattafiori2024llama3herdmodels}
Aaron Grattafiori, Abhimanyu Dubey, Abhinav Jauhri, Abhinav Pandey, Abhishek Kadian, Ahmad Al-Dahle, Aiesha Letman, Akhil Mathur, Alan Schelten, Alex Vaughan, Amy Yang, Angela Fan, Anirudh Goyal, Anthony Hartshorn, Aobo Yang, Archi Mitra, Archie Sravankumar, Artem Korenev, Arthur Hinsvark, and 542 others. 2024.
\newblock \href {https://arxiv.org/abs/2407.21783} {The llama 3 herd of models}.
\newblock \emph{Preprint}, arXiv:2407.21783.

\bibitem[{Harabor and Grastien(2014)}]{10.5555/3038794.3038810}
Daniel Harabor and Alban Grastien. 2014.
\newblock Improving jump point search.
\newblock In \emph{Proceedings of the Twenty-Fourth International Conferenc on International Conference on Automated Planning and Scheduling}, ICAPS'14, page 128–135. AAAI Press.

\bibitem[{Hart et~al.(1968)Hart, Nilsson, and Raphael}]{Hart1968}
Peter Hart, Nils Nilsson, and Bertram Raphael. 1968.
\newblock \href {https://doi.org/10.1109/tssc.1968.300136} {A formal basis for the heuristic determination of minimum cost paths}.
\newblock \emph{{IEEE} Transactions on Systems Science and Cybernetics}, 4(2):100--107.

\bibitem[{Latif(2024)}]{latif20243p}
Ehsan Latif. 2024.
\newblock 3p-llm: Probabilistic path planning using large language model for autonomous robot navigation.
\newblock \emph{arXiv preprint arXiv:2403.18778}.

\bibitem[{Li et~al.(2025)Li, Zhu, Cartis, Ji, and Liu}]{li2025sos1o1r1likereasoning}
Kechen Li, Wenqi Zhu, Coralia Cartis, Tianbo Ji, and Shiwei Liu. 2025.
\newblock \href {https://arxiv.org/abs/2502.20545} {Sos1: O1 and r1-like reasoning llms are sum-of-square solvers}.
\newblock \emph{Preprint}, arXiv:2502.20545.

\bibitem[{Liu et~al.(2023)Liu, Jiang, Zhang, Liu, Zhang, Biswas, and Stone}]{liu2023llm+}
Bo~Liu, Yuqian Jiang, Xiaohan Zhang, Qiang Liu, Shiqi Zhang, Joydeep Biswas, and Peter Stone. 2023.
\newblock Llm+ p: Empowering large language models with optimal planning proficiency.
\newblock \emph{arXiv preprint arXiv:2304.11477}.

\bibitem[{Meng(2025)}]{meng2025llm}
Silin Meng. 2025.
\newblock Llm-a*: Large language model enhanced incremental heuristic search on path planning.
\newblock Master's thesis, University of California, Los Angeles.

\bibitem[{OpenAI et~al.(2024)OpenAI, Achiam, Adler, Agarwal, Ahmad, Akkaya, Aleman, Almeida, Altenschmidt, Altman, Anadkat, Avila, Babuschkin, Balaji, Balcom, Baltescu, Bao, Bavarian, Belgum, Bello, Berdine, Bernadett-Shapiro, Berner, Bogdonoff, Boiko, Boyd, Brakman, Brockman, Brooks, Brundage, Button, Cai, Campbell, Cann, Carey, Carlson, Carmichael, Chan, Chang, Chantzis, Chen, Chen, Chen, Chen, Chen, Chess, Cho, Chu, Chung, Cummings, Currier, Dai, Decareaux, Degry, Deutsch, Deville, Dhar, Dohan, Dowling, Dunning, Ecoffet, Eleti, Eloundou, Farhi, Fedus, Felix, Fishman, Forte, Fulford, Gao, Georges, Gibson, Goel, Gogineni, Goh, Gontijo-Lopes, Gordon, Grafstein, Gray, Greene, Gross, Gu, Guo, Hallacy, Han, Harris, He, Heaton, Heidecke, Hesse, Hickey, Hickey, Hoeschele, Houghton, Hsu, Hu, Hu, Huizinga, Jain, Jain, Jang, Jiang, Jiang, Jin, Jin, Jomoto, Jonn, Jun, Kaftan, Łukasz Kaiser, Kamali, Kanitscheider, Keskar, Khan, Kilpatrick, Kim, Kim, Kim, Kirchner, Kiros, Knight, Kokotajlo, Łukasz Kondraciuk,
  Kondrich, Konstantinidis, Kosic, Krueger, Kuo, Lampe, Lan, Lee, Leike, Leung, Levy, Li, Lim, Lin, Lin, Litwin, Lopez, Lowe, Lue, Makanju, Malfacini, Manning, Markov, Markovski, Martin, Mayer, Mayne, McGrew, McKinney, McLeavey, McMillan, McNeil, Medina, Mehta, Menick, Metz, Mishchenko, Mishkin, Monaco, Morikawa, Mossing, Mu, Murati, Murk, Mély, Nair, Nakano, Nayak, Neelakantan, Ngo, Noh, Ouyang, O'Keefe, Pachocki, Paino, Palermo, Pantuliano, Parascandolo, Parish, Parparita, Passos, Pavlov, Peng, Perelman, de~Avila Belbute~Peres, Petrov, de~Oliveira~Pinto, Michael, Pokorny, Pokrass, Pong, Powell, Power, Power, Proehl, Puri, Radford, Rae, Ramesh, Raymond, Real, Rimbach, Ross, Rotsted, Roussez, Ryder, Saltarelli, Sanders, Santurkar, Sastry, Schmidt, Schnurr, Schulman, Selsam, Sheppard, Sherbakov, Shieh, Shoker, Shyam, Sidor, Sigler, Simens, Sitkin, Slama, Sohl, Sokolowsky, Song, Staudacher, Such, Summers, Sutskever, Tang, Tezak, Thompson, Tillet, Tootoonchian, Tseng, Tuggle, Turley, Tworek, Uribe, Vallone,
  Vijayvergiya, Voss, Wainwright, Wang, Wang, Wang, Ward, Wei, Weinmann, Welihinda, Welinder, Weng, Weng, Wiethoff, Willner, Winter, Wolrich, Wong, Workman, Wu, Wu, Wu, Xiao, Xu, Yoo, Yu, Yuan, Zaremba, Zellers, Zhang, Zhang, Zhao, Zheng, Zhuang, Zhuk, and Zoph}]{openai2024gpt4technicalreport}
OpenAI, Josh Achiam, Steven Adler, Sandhini Agarwal, Lama Ahmad, Ilge Akkaya, Florencia~Leoni Aleman, Diogo Almeida, Janko Altenschmidt, Sam Altman, Shyamal Anadkat, Red Avila, Igor Babuschkin, Suchir Balaji, Valerie Balcom, Paul Baltescu, Haiming Bao, Mohammad Bavarian, Jeff Belgum, and 262 others. 2024.
\newblock \href {https://arxiv.org/abs/2303.08774} {Gpt-4 technical report}.
\newblock \emph{Preprint}, arXiv:2303.08774.

\bibitem[{Qwen et~al.(2025)Qwen, :, Yang, Yang, Zhang, Hui, Zheng, Yu, Li, Liu, Huang, Wei, Lin, Yang, Tu, Zhang, Yang, Yang, Zhou, Lin, Dang, Lu, Bao, Yang, Yu, Li, Xue, Zhang, Zhu, Men, Lin, Li, Tang, Xia, Ren, Ren, Fan, Su, Zhang, Wan, Liu, Cui, Zhang, and Qiu}]{qwen2025qwen25technicalreport}
Qwen, :, An~Yang, Baosong Yang, Beichen Zhang, Binyuan Hui, Bo~Zheng, Bowen Yu, Chengyuan Li, Dayiheng Liu, Fei Huang, Haoran Wei, Huan Lin, Jian Yang, Jianhong Tu, Jianwei Zhang, Jianxin Yang, Jiaxi Yang, Jingren Zhou, and 25 others. 2025.
\newblock \href {https://arxiv.org/abs/2412.15115} {Qwen2.5 technical report}.
\newblock \emph{Preprint}, arXiv:2412.15115.

\bibitem[{Valmeekam et~al.(2023)Valmeekam, Marquez, Sreedharan, and Kambhampati}]{valmeekam2023planningabilitieslargelanguage}
Karthik Valmeekam, Matthew Marquez, Sarath Sreedharan, and Subbarao Kambhampati. 2023.
\newblock \href {https://arxiv.org/abs/2305.15771} {On the planning abilities of large language models : A critical investigation}.
\newblock \emph{Preprint}, arXiv:2305.15771.

\bibitem[{Wang et~al.(2023)Wang, Lyu, Ji, Zhang, Yu, Shi, and Tu}]{wang2023documentlevelmachinetranslationlarge}
Longyue Wang, Chenyang Lyu, Tianbo Ji, Zhirui Zhang, Dian Yu, Shuming Shi, and Zhaopeng Tu. 2023.
\newblock \href {https://arxiv.org/abs/2304.02210} {Document-level machine translation with large language models}.
\newblock \emph{Preprint}, arXiv:2304.02210.

\bibitem[{Wei et~al.(2023)Wei, Wang, Schuurmans, Bosma, Ichter, Xia, Chi, Le, and Zhou}]{chainofthoughtpromptingelicitsreasoning}
Jason Wei, Xuezhi Wang, Dale Schuurmans, Maarten Bosma, Brian Ichter, Fei Xia, Ed~Chi, Quoc Le, and Denny Zhou. 2023.
\newblock \href {https://arxiv.org/abs/2201.11903} {Chain-of-thought prompting elicits reasoning in large language models}.
\newblock \emph{Preprint}, arXiv:2201.11903.

\bibitem[{Wen(2025)}]{wen2025language}
Ximing Wen. 2025.
\newblock Language model meets prototypes: Towards interpretable text classification models through prototypical networks.
\newblock In \emph{Proceedings of the AAAI Conference on Artificial Intelligence}, volume~39, pages 29307--29308.

\bibitem[{Yao et~al.(2023)Yao, Zhao, Yu, Du, Shafran, Narasimhan, and Cao}]{yao2023react}
Shunyu Yao, Jeffrey Zhao, Dian Yu, Nan Du, Izhak Shafran, Karthik Narasimhan, and Yuan Cao. 2023.
\newblock React: Synergizing reasoning and acting in language models.
\newblock In \emph{International Conference on Learning Representations (ICLR)}.

\end{thebibliography}

\appendix
\clearpage
\onecolumn
\appendix

\section{Prompt Design}
\label{appendix:01}
\subsection{Independent Route Planning Prompts}
\begin{table}[h]
    \vspace{-1em}
    \centering
    \fontsize{10pt}{12pt}\selectfont
    \fbox{
        \begin{minipage}{0.95\textwidth}
        Given these obstacle coordinates in a grid\{obstacles\}:
        \\
        Please independently plan a continuous path of coordinates from the starting point (\{start\_x\}, \{start\_y\})to the ending point (\{end\_x\}, \{end\_y\}), avoiding all obstacles listed above. \\
        The path must follow these rules:\\
        1. Every point must differ by exactly 1 in either x or y direction from the previous point (no diagonal moves allowed).\\
        2. The output should be in the format: [(x1, y1), (x2, y2), (x3, y3), ...].\\
        3. If a valid path cannot be created without touching an obstacle, return an empty path [].\\
        Return only the path without any additional explanation.
        \end{minipage}
     }
    \centering
    \caption{Vanilla prompt}
    
\end{table}

% \begin{table}[h]
% \centering
% \caption{Vanilla Prompt}\label{fig:Vanilla}
% \begin{tabularx}{\textwidth}{|X|}
%     \hline
%     Given these obstacle coordinates in a grid\{obstacles\}:\bigstrut[t]\\
%     Please independently plan a continuous path of coordinates from the starting point (\{start\_x\}, \{start\_y\})to the ending point (\{end\_x\}, \{end\_y\}), avoiding all obstacles listed above.\\
%     The path must follow these rules:\\
%     1. Every point must differ by exactly 1 in either x or y direction from the previous point (no diagonal moves allowed).\\
%     2. The output should be in the format: [(x1, y1), (x2, y2), (x3, y3), ...].\\
%     3. If a valid path cannot be created without touching an obstacle, return an empty path [].\\
%     Return only the path without any additional explanation.\bigstrut[b]\\
%     \hline
% \end{tabularx}
% \end{table}

\begin{table}[h]
    \vspace{-1.5em}
    \centering
    \fontsize{10pt}{12pt}\selectfont
    \fbox{
        \begin{minipage}{0.95\textwidth}
        Here are the obstacle coordinates in the grid:\{obstacles\}"\\
        You need to plan a path from the starting point (\{start\_x\}, \{start\_y\}) to the ending point (\{end\_x, end\_y) while avoiding the obstacles. \\
        To ensure accuracy, follow these steps step by step:\\
        1. Verify that the starting point \{(start\_x\},\{start\_y)\} and ending point (\{end\_x, \{end\_y\}) are not obstacles. \\
        2. List all valid moves for a point in the grid. A move is valid if: \\
            - It differs by exactly 1 in either the x or y direction from the previous point. \\
            - It does not overlap with any obstacle. \\
        3. Starting from (\{start\_x\}, \{start\_y\}), iteratively choose the next valid move to build a path to (\{end\_x\}, \{end\_y\}). \\
        4. If there is no valid path, return an empty path []. \\
        Return the final path in this format: [(x1, y1), (x2, y2), (x3, y3), ...]. If there are any errors or no valid paths, return []. \\
        Make sure to provide only the path in your response without any extra explanations. 
        \end{minipage}
     }
    \centering
    \caption{CoT prompt}
    \label{fig:CoT}
\end{table}

% \begin{table}[h]
% \centering
% \caption{CoT Prompt}\label{fig:CoT}
% \begin{tabularx}{\textwidth}{|X|}
%     \hline
%     Here are the obstacle coordinates in the grid:"\{obstacles\}"\bigstrut[t]\\
%     You need to plan a path from the starting point (\{start\_x\}, \{start\_y\}) to the ending point (\{end\_x, end\_y) while avoiding the obstacles. \\
%     To ensure accuracy, follow these steps step by step:\\
%     1. Verify that the starting point \{(start\_x\},\{start\_y)\} and ending point (\{end\_x, \{end\_y\}) are not obstacles. \\
%     2. List all valid moves for a point in the grid. A move is valid if: \\
%     - It differs by exactly 1 in either the x or y direction from the previous point. \\
%     - It does not overlap with any obstacle. \\
%     3. Starting from (\{start\_x\}, \{start\_y\}), iteratively choose the next valid move to build a path to (\{end\_x\}, \{end\_y\}). \\
%     4. If there is no valid path, return an empty path []. \\
%     Return the final path in this format: [(x1, y1), (x2, y2), (x3, y3), ...]. If there are any errors or no valid paths, return []. \\
%     Make sure to provide only the path in your response without any extra explanations.\bigstrut[b]\\
%     \hline
% \end{tabularx}
% \end{table}

\begin{table}[h!]
    \centering
    \fontsize{10pt}{12pt}\selectfont
    \fbox{
        \begin{minipage}{0.95\textwidth}
        Plan the shortest path from the starting point (\{start\_x\}, \{start\_y\}) to the endpoint (\{end\_x\}, \{end\_y\}) while avoiding rectangular obstacles defined as \{obstacles\}. Follow these steps:
        \vspace{0.3cm}
        \\
        1. Analyze the grid to identify the start, end, and rectangular obstacle locations.\\
        2. Each rectangle is defined by its top-left corner (x1,  y1) and bottom-right corner (x2, y2). A grid cell (x, y) is an obstacle if $x1 \leq x \leq x2$ and $y1 \leq y \leq y2$.\\
        3. Define valid moves: \\
            \hspace*{1em} -Up:\ $(x, y+1)$ \\
            \hspace*{1em} -Down:\ $(x, y-1)$ \\
            \hspace*{1em} -Right:\ $(x+1, y)$ \\
            \hspace*{1em} -Left:\ $(x-1, y)$ \\
        4. Build the path:\\
            \hspace*{1em} - Start at (\{start\_x\}, \{start\_y\}).\\
            \hspace*{1em} - At each step, evaluate valid moves and select the one that minimizes the distance to (\{end\_x\}, \{end\_y\}).\\
        5. If no valid moves are available, return an empty path [].\\
        6. Output the result in this format: [(x1, y1), (x2, y2), ..., (xn, yn)].\\
        Example 1:\\
        Start: (3, 7), End: (4, 3), Obstacles: [((1, 2), (3, 4)), ((2, 5), (4, 6))]\\
        Path: [(3, 7), (4, 7), (5, 7), (5, 6), (5, 5), (5, 4), (5, 3), (4, 3)]\\
        Example 2:\\
        Start: (2, 4), End: (7, 5), Obstacles: [((3, 4), (5, 6))]\\
        Path: [(2, 4), (2, 3), (3, 3), (4, 3), (5, 3), (6, 3), (7, 3), (7, 4), (7, 5)]\\
        Return only the path as output, without additional explanations.
        \end{minipage}
     }
    \centering
    \caption{FewShot-Base prompt}
    \label{fig:Few-shot Learning prompt}
\end{table}

\clearpage
\subsection{Algorithm of Planning(AoP) Prompts}
\begin{table}[h!]
    \centering
    \fontsize{10pt}{12pt}\selectfont
    \fbox{
        \begin{minipage}{0.95\textwidth}
        Plan a path from the starting point (\{start\_x\}, \{start\_y\}) to the endpoint (\{end\_x\}, \{end\_y\}) while avoiding rectangular obstacles defined as \{obstacles\}. Use DFS algorithm to explore possible paths. Follow these steps:\\
        \\
        1. Analyze the grid to identify the start, end, and rectangular obstacle locations.\\
        \\
        2. Each rectangle is defined by its top-left corner (x1, y1) and bottom-right corner (x2, y2). A grid cell (x, y) is an obstacle if $x1 \leq x \leq x2$ and $y1 \leq y \leq y2$.\\ 
        \\
        3. Define valid moves (same priority order):\\
        \hspace*{1em} - Up: (x, y+1)\\
        \hspace*{1em} - Down: (x, y-1)\\
        \hspace*{1em} - Right: (x+1, y)\\
        \hspace*{1em} - Left: (x-1, y)\\
        \hspace*{1em} - All moves must stay within the grid bounds and avoid obstacle cells.\\ 
        \\
        4. Use Depth-First Search (DFS) algorithm:\\
        \hspace*{1em} - Initialize a stack with the start cell and its path [(start\_x, start\_y)]\\
        \hspace*{1em} - Create a visited set to track explored cells\\
        \hspace*{1em} - While stack is not empty:\\
        \hspace*{1em} - Pop the last cell and its path from the stack\\
        \hspace*{1em} - If current cell is the endpoint, return the complete path\\
        \hspace*{1em} - If cell not in visited:\\
        \hspace*{2em} - Mark cell as visited\\
        \hspace*{2em} - Push all valid neighboring cells to the stack (FILO order):\\
        \hspace*{2em} * For each neighbor in move order (Right→Left→Down→Up):\\
        \hspace*{3em} - If neighbor is within grid bounds\\
        \hspace*{3em} - If neighbor is not an obstacle\\
        \hspace*{3em} - If neighbor not in visited\\
        \hspace*{3em} - Push (neighbor\_coords, path + [neighbor\_coords]) to stack\\
        \\
        5. Termination conditions:\\
        \hspace*{1em}   - If stack becomes empty before reaching endpoint, return empty path [].\\
        \hspace*{1em}   - First valid path found using DFS should be returned.\\
        \\
        6. Output the result in this format: [(x1, y1), (x2, y2), ..., (xn, yn)].\\
        \\
        Return only the path as output, without additional explanations.\\
 \end{minipage}
     }
    \centering
    \caption{AoP-DFS prompt}
    \label{fig:AoT(DFS) prompt}
\end{table}

\clearpage
\begin{table}[htb!]
    \vspace{-1.5em}
    \centering
    \fontsize{10pt}{12pt}\selectfont
    \fbox{
        \begin{minipage}{0.95\textwidth}
        Plan the shortest path from the starting point (\{start\_x\}, \{start\_y\}) to the endpoint (\{end\_x\}, \{end\_y\}) while avoiding rectangular obstacles defined as \{obstacles\}. Use A* algorithm to calculate the shortest path. Follow these steps:\\
         \\
        1. Analyze the grid to identify the start, end, and rectangular obstacle locations.\\
         \\
        2. Each rectangle is defined by its top-left corner (x1, y1) and bottom-right corner (x2, y2). A grid cell (x, y) is an obstacle if $x1 \leq x \leq x2$ and $y1 \leq y \leq y2$.\\
         \\
        3. Define valid moves:\\
        \hspace*{1em} - Up: (x, y+1)\\
        \hspace*{1em} - Down: (x, y-1)\\
        \hspace*{1em} - Right: (x+1, y)\\
        \hspace*{1em} - Left: (x-1, y)\\
        \hspace*{1em} - All moves must stay within the grid bounds and avoid obstacle cells.\\
         \\
        4. Use A* algorithm:\\
        \hspace*{1em} - Initialize g\_score (cost from start) for all cells to infinity, except the start cell which is 0.\\
        \hspace*{1em} - Initialize f\_score (g\_score + heuristic) for all cells to infinity, except the start cell which is heuristic(start, end).\\
        \hspace*{1em} - Use a priority queue to repeatedly select the grid cell with the lowest f\_score.\\
        \hspace*{1em} - For each selected cell, evaluate its valid neighbors:\\
        \hspace*{1em} - Calculate the tentative g\_score for the neighbor through the current cell.\\
        \hspace*{1em} - If the tentative g\_score is lower than the neighbor's current g\_score, update the neighbor's parent, g\_score, and f\_score (g\_score + heuristic(neighbor, end)).\\
        \hspace*{1em} - Add the neighbor to the priority queue if it's not already there or if this path is better.\\
        \hspace*{1em} - Use the Manhattan distance as the heuristic function: h((x1, y1), (x2, y2)) = abs(x1 - x2) + abs(y1 - y2).\\
        \hspace*{1em} - Continue until the endpoint ({end\_x}, {end\_y}) is reached or the priority queue is empty.\\
        \\
        5. Construct the path:\\
        \hspace*{1em} - Start at ({start\_x}, {start\_y}).\\
        \hspace*{1em} - Trace back from the endpoint to the starting point using the recorded parent cells to reconstruct the shortest path.\\
        \hspace*{1em} - If no valid moves are available, return an empty path [].\\
        \\
        6. Output the result in this format: [(x1, y1), (x2, y2), ..., (xn, yn)].\\
        Return only the path as output, without additional explanations.\\
    \end{minipage}
     }
    \centering
    \caption{AoP-A* prompt}
    \label{fig:AoT(A*)}
\end{table}

\clearpage
\begin{table}[htb!]
    \centering
    \vspace{-1.5em}
    \fontsize{10pt}{12pt}\selectfont
    \fbox{
        \begin{minipage}{0.95\textwidth}
        Plan the shortest path from the starting point (\{start\_x\}, \{start\_y\}) to the endpoint (\{end\_x\}, \{end\_y\}) while avoiding rectangular obstacles defined as \{obstacles\}. Use Dijkstra's algorithm to calculate the shortest path. Follow these steps:\\
        \\
        1. Analyze the grid to identify the start, end, and rectangular obstacle locations.\\    
        \\
        2. Each rectangle is defined by its top-left corner (x1, y1) and bottom-right corner (x2, y2). A grid cell (x, y) is an obstacle if $x1 \leq x \leq x2$ and $y1 \leq y \leq y2$.\\   
        \\
        3. Define valid moves:\\        
        \hspace*{1em}   - Up: (x, y+1)\\     
        \hspace*{1em}   - Down: (x, y-1)\\      
        \hspace*{1em}   - Right: (x+1, y)\\        
        \hspace*{1em}   - Left: (x-1, y)\\       
        \hspace*{1em}   - All moves must stay within the grid bounds and avoid obstacle cells.\\
        \\
        4. Use Dijkstra's algorithm:\\
        \hspace*{1em}   - Assign an initial distance of infinity to all grid cells except the starting point, which should have a distance of 0.\\  
        \hspace*{1em}   - Use a priority queue to repeatedly select the grid cell with the smallest tentative distance.\\
        \hspace*{1em}   - For each selected cell, evaluate its valid neighbors and update their tentative distances if a shorter path is found.\\
        \hspace*{1em}   - Continue until the endpoint (\{end\_x\}, \{end\_y\}) is reached or all possible paths have been explored.\\
        \\
        5. Construct the path:"\\
        \hspace*{1em}   - Start at (\{start\_x\}, \{start\_y\}).\\
        \hspace*{1em}   - Trace back from the endpoint to the starting point using the recorded parent cells to reconstruct the shortest path.\\
        \hspace*{1em}   - If no valid moves are available, return an empty path [].\\ 
        \\
        6. Output the result in this format: [(x1, y1), (x2, y2), ..., (xn, yn)].\\
        Return only the path as output, without additional explanations.\\
 \end{minipage}
     }
    \centering
    \caption{AoP-Dijkstra prompt}
    \label{fig:AoT(Dijkstra)}
\end{table}

\clearpage
\subsection{AoP with Example (Algo-Shot) Prompt}
\begin{table}[htb!]
    \centering
    \fontsize{10pt}{12pt}\selectfont
    \fbox{
        \begin{minipage}{0.95\textwidth}
        Plan the shortest path from the starting point (\{start\_x\}, \{start\_y\}) to the endpoint (\{end\_x\}, \{end\_y\}) while avoiding rectangular obstacles defined as \{obstacles\}. Use Dijkstra's algorithm to calculate the shortest path. Follow these steps:"\\
        \\
        1. Analyze the grid to identify the start, end, and rectangular obstacle locations.\\  
        \\
        2. Each rectangle is defined by its top-left corner (x1, y1) and bottom-right corner (x2, y2). A grid cell (x, y) is an obstacle if $x1 \leq x \leq x2$ and $y1 \leq y \leq y2$.\\   
        \\
        3. Define valid moves:\\        
        \hspace*{1em}   - Up: (x, y+1)\\     
        \hspace*{1em}   - Down: (x, y-1)\\      
        \hspace*{1em}   - Right: (x+1, y)\\        
        \hspace*{1em}   - Left: (x-1, y)\\       
        \hspace*{1em}   - All moves must stay within the grid bounds and avoid obstacle cells.\\
        \\
        4. Use Dijkstra's algorithm:\\
        \hspace*{1em}   - Assign an initial distance of infinity to all grid cells except the starting point, which should have a distance of 0.\\  
        \hspace*{1em}   - Use a priority queue to repeatedly select the grid cell with the smallest tentative distance.\\
        \hspace*{1em}   - For each selected cell, evaluate its valid neighbors and update their tentative distances if a shorter path is found.\\
        \hspace*{1em}   - Continue until the endpoint (\{end\_x\}, \{end\_y\}) is reached or all possible paths have been explored.\\
        \\
        5. Construct the path:"\\
        \hspace*{1em}   - Start at (\{start\_x\}, \{start\_y\}).\\
        \hspace*{1em}   - Trace back from the endpoint to the starting point using the recorded parent cells to reconstruct the shortest path.\\
        \hspace*{1em}   - If no valid moves are available, return an empty path [].\\ 
        \\
        6. Output the result in this format: [(x1, y1), (x2, y2), ..., (xn, yn)].\\ 
        \\
        Example 1:\\
        Start: (3, 7), End: (4, 3), Obstacles: [((1, 2), (3, 4)), ((2, 5), (4, 6))]\\
        Path: [(3, 7), (4, 7), (5, 7), (5, 6), (5, 5), (5, 4), (5, 3), (4, 3)]\\
        \\
        Example 2:\\
        Start: (2, 4), End: (7, 5), Obstacles: [((3, 4), (5, 6))]\\
        Path: [(2, 4), (2, 3), (3, 3), (4, 3), (5, 3), (6, 3), (7, 3), (7, 4), (7, 5)]\\\
        \\
        Return only the path as output, without additional explanations.\\
 \end{minipage}
     }
    \caption{\centering Algo-Direct prompt}
    \label{fig:Easy shot prompt}
\end{table}

\clearpage
\begin{table}[h!]
    \vspace{-2.0em}
    \centering
    \fontsize{8pt}{10pt}\selectfont
    \newgeometry{margin=0.8in}
    \fbox{
        \begin{minipage}{0.95\textwidth}
        Here are the obstacle coordinates in the grid:\{obstacles\}\\
        You need to plan a path from the starting point (\{start\_x\}, \{start\_y\}) to the ending point (\{end\_x\}, \{end\_y\}) while avoiding the obstacles. \\
        Use Dijkstra's algorithm principles to ensure the path is valid and follows the shortest possible route. Follow these steps:\\
        1. Verify the starting point (\{start\_x\}, \{start\_y\}) and ending point (\{end\_x\}, \{end\_y\}) are not obstacles.\\
        2. Define the cost of reaching a point in the grid:\\
        \hspace*{1em} - Start by assigning a cost of 0 to the starting point (\{start\_x\}, \{start\_y\}).\\
        \hspace*{1em} - For all other points, initialize the cost as infinity ($\infty$).\\
        3. Use a priority queue to iteratively explore points in the grid:\\
        \hspace*{1em} - Begin with the starting point (\{start\_x\}, \{start\_y\}) in the queue.
        \hspace*{1em} - At each step, select the point with the lowest current cost.\\
        \hspace*{1em} - For each valid move (up, down, left, right):\\
        \hspace*{2em} a. If moving to a neighboring point reduces its cost, update the cost and add the point to the queue.\\
        \hspace*{2em} b. A move is valid if it does not overlap with any obstacle and stays within the grid.\\
        4. Repeat this process until reaching the ending point (\{end\_x\}, \{end\_y\}) or exhausting all valid points in the queue:\\
        \hspace*{1em} - If the ending point is reached, reconstruct the path by tracing back from the endpoint to the start using the cost values.\\
        \hspace*{1em} - If no valid path exists, return an empty path [].\\
        5. Return the final path in this format: [(x1, y1), (x2, y2), (x3, y3), ...]. If there are any errors or no valid paths, return [].\\
        
        Example:\\
        Start: (3, 7), End: (4, 3), Obstacles: [((1, 2), (3, 4)), ((2, 5), (4, 6))], grid: 10\\
        Evaluate valid neighbors in given map:\\
        \hspace*{1em} - Starting at (3, 7), distance is 0 :\\
        \hspace*{1em} - (3, 6): Obstructed, skip.\\
        \hspace*{1em} - (4, 7): Valid, update distance to 1.\\
        \hspace*{1em} - (2, 7): Valid, update distance to 1.\\
        \hspace*{1em} - (3, 8): Valid, update distance to 1.\\
        
        \hspace*{1em} - To (3, 6): Obstructed point.\\     
        
        \hspace*{1em} - To (3, 8), recorded distance is 1:\\   
        \hspace*{1em} - (3, 7): Start point, distance is 0.\\         
        \hspace*{1em} - (4, 8): Valid, update distance to 2.\\
        \hspace*{1em} - (2, 8): Valid, update distance to 2.\\
        \hspace*{1em} - (3, 9): Valid, update distance to 2.\\
        
        \hspace*{1em} - To (3, 9), recorded distance is 2:\\
        \hspace*{1em} - (3, 8):  Already visited, distance is,3, but recorded distance is $1 < 3$, so distance stays at 1.\\
        \hspace*{1em}  - (4, 9): Valid, update distance to 3.\\
        \hspace*{1em}  - (2, 9): Valid, update distance to 3.\\
        \hspace*{1em}  - (3, 10): Out of grids, skip.\\
        
        \hspace*{1em}  - To (4, 9), recorded distance is 3:\\       
        \hspace*{1em}  - (4, 8): Already visited, distance is 4, but recorded distance is $2 < 4$, so distance stays at 2.\\
        \hspace*{1em}  - (5, 9): Valid, update distance to 4.\\
        \hspace*{1em}  - (3, 9): Already visited, distance is 4, but recorded distance is $2 < 4$, so distance stays at 2.\\
        \hspace*{1em}  - (4, 10): Out of grids, skip\\
        
        \hspace*{1em} - To (5, 9), recorded distance is 4:\\
        \hspace*{1em} - (5, 8): Valid, update distance to 5.\\
        \hspace*{1em} - (6, 9): Valid, update distance to 5.\\
        \hspace*{1em} - (4, 9): Already visited, distance is 5, but recorded distance is $3 < 4$, so distance stays at 3.\\
        \hspace*{1em}  - (5, 10): Out of grids, skip.
        
        \hspace*{1em} - To (4, 8), recorded distance is 2:\\
        \hspace*{1em} - (4, 7): Already visited, distance is 3, but recorded distance is 1 < 3, so distance stays at 1.\\
        \hspace*{1em}  - (5, 8): Already visited, distance is 3, but recorded distance is 5 > 3, so distance updates to 3.\\
        \hspace*{1em}  - (3, 8): Already visited, distance is 3, but recorded distance is 1 < 3, so distance stays at 1.\\
        \hspace*{1em}  - (4, 9): Already visited, distance is 3 which is equal to recorded distance, so distance stays at 3.\\
        
        \hspace*{1em}  - Traverse all point in the map\\
        
        \hspace*{1em} - Result: the distance of (4,3) is 7, which means (3,7) to (4,3) need 7 steps.\\
        
        \hspace*{1em} - Generated Path: [(3, 7), (4, 7), (5, 7), (5, 6), (5, 5), (5, 4), (4, 4), (4, 3)]\\
        
        Now, apply the same logic to plan a path from (\{start\_x\}, \{start\_y\}) to (\{end\_x\}, \{end\_y\}) while avoiding the obstacles:\\
            
        Make sure to provide only the path in your response without any extra explanations.
 \end{minipage}
     }
    \caption{\centering Algo-Reasoning prompt}
    \label{fig:Complex shot}
\end{table}

\section{Results}
\label{appendix:02}

\begin{table*}[htbp]
\centering
\vspace*{-5.0em}
\label{tab:performance of Native Plan}
\scriptsize % 使用更小的字体
\setlength{\tabcolsep}{3pt} % 减少列间距
\begin{tabular}{@{}lc*{3}{cccccc}@{}}
\toprule
\multirow{2}{*}{Model} & \multirow{2}{*}{Prompt} & \multicolumn{6}{c}{10 size} & \multicolumn{6}{c}{20 size} & \multicolumn{6}{c}{30 size} \\
\cmidrule(lr){3-8} \cmidrule(lr){9-14} \cmidrule(lr){15-20}
 & & \multicolumn{1}{c}{CR} & \multicolumn{1}{c}{FR} & \multicolumn{1}{c}{OR} & \multicolumn{1}{c}{GM} & \multicolumn{1}{c}{MSE} & \multicolumn{1}{c}{RT(s)} & 
 \multicolumn{1}{c}{CR} & \multicolumn{1}{c}{FR} & \multicolumn{1}{c}{OR} & \multicolumn{1}{c}{GM} & \multicolumn{1}{c}{MSE} & \multicolumn{1}{c}{RT(s)} & 
 \multicolumn{1}{c}{CR} & \multicolumn{1}{c}{FR} & \multicolumn{1}{c}{OR} & \multicolumn{1}{c}{GM} & \multicolumn{1}{c}{MSE} & \multicolumn{1}{c}{RT(s)} \\
\midrule
deepseek-V3 & Vanilla & 67 & 48 & 47 & 104.1 & 3 & 1.9 & 37 & 26 & 26 & 100 & 0 & 2.4 & 19 & 11 & 10 & 109.2 & 29.5 & 3.0 \\
qwen2.5-7b-instruct & & 100 & 48 & 47 & 101.5 & 8.3 & 2.4 & 99 & 25 & 24 & 101.4 & 1.4 & 12.5 & 99 & 13 & 13 & 100 & 0 & 35.1 \\
qwen2.5-32b-instruct & & 98 & 63 & 45 & 121.1 & 12.9 & 2.2 & 99 & 52 & 48 & 104.8 & 4.9 & 4.3 & 98 & 64 & 61 & 104.4 & 17.4 & 7.0 \\
qwen2.5-72b-instruct & & 100 & 72 & 69 & 104.0 & 3.3 & 1.9 & 100 & 61 & 58 & 105.3 & 16.7 & 3.7 & 100 & 57 & 54 & 105.4 & 29.8 & 5.9 \\
llama3-70b-instruct & & 100 & 69 & 52 & 124.5 & 15.2 & 1.2 & 100 & 57 & 48 & 112.5 & 35.1 & 1.9 & 100 & 54 & 48 & 113.9 & 93.3 & 2.8 \\
ChatGPT-4 & & 100 & 76 & 63 & 112.2 & 11.5 & 2.3 & 99 & 61 & 48 & 113.5 & 36.8 & 3.7 & 98 & 62 & 52 & 112.2 & 101.0 & 5.1 \\
\midrule
deepseek-V3 & CoT & 100 & 73  & 72  & 100.39 & 0.054 & 2.95 & 93 & 68 & 68 & 100.00 & 0 & 5.54 & 76 & 53 & 53 & 100.00 & 0 & 7.54 \\
qwen2.5-7b-instruct & & 100 & 64 & 58 & 106.12 & 2.18 & 1.89 & 100 & 66 & 64 & 101.57 & 1.03 & 3.94 & 93 & 54 & 50 & 104.26 & 12.96 & 6.01 \\
qwen2.5-32b-instruct & & 100 & 63 & 57 & 105.54 & 2.03 & 2.21 & 100 & 67 & 66 & 100.50 & 0.05 & 4.03 & 94 & 54 & 51 & 103.03 & 6.96 & 6.27 \\   
qwen2.5-72b-instruct & & 100 & 64 & 58 & 106.12 & 2.18 & 1.89 & 100 & 66 & 64 & 101.57 & 1.03 & 3.94 & 93 & 54 & 50 & 104.26 & 12.96 & 6.01 \\
llama3-70b-instruct & & 100 & 70 & 64 & 105.86 & 2.91 & 1.24 & 100 & 60 & 60 & 100 & 0 & 1.87 & 100  & 62 & 62 & 100 & 0 & 2.32 \\
ChatGPT-4 & & 100 & 84  & 77 & 102.31 & 0.71 & 2.47 & 100 & 75  & 64 & 104.79 & 21.33 & 4.00 & 100 & 69 & 64 & 103 & 39.13 & 6.25 \\
\midrule
deepseek-V3 & Few-shot & 100  & 85  & 84  & 100.03 & 0.18 & 1.94 & 100 & 70 & 70 & 100 & 0 & 2.61 & 100 & 67 & 67 & 100 & 0 & 3.37 \\
qwen2.5-7b-instruct & Learning & 100 & 44 & 38 & 109.28 & 10.09 & 3.47 & 100 & 37 & 35 & 102.54 & 4.32 & 4.48 & 99 & 21 & 19 & 102.89 & 4.76 & 8.71 \\
qwen2.5-32b-instruct & & 100 & 65 & 55 & 108.18 & 3.81 & 2.05 & 100 & 54 & 46 & 108.67 & 26.37 & 4.62 & 100 & 60 & 42 & 121.79 & 134.03 & 6.85 \\
qwen2.5-72b-instruct & & 100 & 76 & 71 & 102.85 & 2.10 & 2.15 & 100 & 62 & 60 & 101.57 & 2.38 & 4.30 & 100 & 60 & 59 & 100.24 & 106.67 & 4.30 \\
llama3-70b-instruct & & 100 & 70 & 68 & 102.32 & 0.28 & 1.06 & 99 & 56 & 56 & 100 & 0 & 11.57 & 100 & 56 & 56 & 100 & 0 & 9.90 \\
ChatGPT-4 & & 85 & 62  & 62 & 100 & 0 & 2.9 & 98 & 68 & 65 & 101.1 & 1.23 & 4.26 & 93 & 67 & 62 & 100.3 & 38.02 & 6.88 \\
\bottomrule
\end{tabular}
\caption{Evaluation metrics of different models with independent route planning prompts across $10\times10$, $20\times20$, and $30\times30$ maps. Results are grouped by prompt technique and model, with bold indicating top performance per group.}
\end{table*}
\begin{table*}[h]
\centering
\vspace*{-5.0em}
\label{tab:performance of AoT}
\scriptsize % 使用更小的字体
\setlength{\tabcolsep}{3pt} % 减少列间距
\begin{tabular}{@{}lc*{3}{cccccc}@{}}
\toprule
\multirow{2}{*}{Model} & \multirow{2}{*}{Prompt} & \multicolumn{6}{c}{10 size} & \multicolumn{6}{c}{20 size} & \multicolumn{6}{c}{30 size} \\
\cmidrule(lr){3-8} \cmidrule(lr){9-14} \cmidrule(lr){15-20}
 & & \multicolumn{1}{c}{CR} & \multicolumn{1}{c}{FR} & \multicolumn{1}{c}{OR} & \multicolumn{1}{c}{GM} & \multicolumn{1}{c}{MSE} & \multicolumn{1}{c}{RT(s)} & 
 \multicolumn{1}{c}{CR} & \multicolumn{1}{c}{FR} & \multicolumn{1}{c}{OR} & \multicolumn{1}{c}{GM} & \multicolumn{1}{c}{MSE} & \multicolumn{1}{c}{RT(s)} & 
 \multicolumn{1}{c}{CR} & \multicolumn{1}{c}{FR} & \multicolumn{1}{c}{OR} & \multicolumn{1}{c}{GM} & \multicolumn{1}{c}{MSE} & \multicolumn{1}{c}{RT(s)} \\
\midrule
deepseek-V3 & AoP & 100 & 79 & 65 & 111.64 & 5.87 & 8.38 & 100 & 67 & 62 & 105.28 & 25.43 & 10.26 & 100 & 72 & 59 & 114.34 & 131.77 & 14.40\\
qwen2.5-7b-instruct & DFS & 100 & 51 & 47 & 105.68 & 2.11 & 2.59 & 100 & 27 & 26 & 101.19 & 2.37 & 3.88 & 100 & 19 & 15 & 106.26 & 9.89 & 12.39\\
qwen2.5-32b-instruct & & 100 & 62 & 49 & 116.14 & 4.77 & 3.09 & 100 & 64 & 50 & 118.79 & 23.375 & 4.90 & 97 & 52 & 43 & 112.98 & 112.85 & 8.00\\
qwen2.5-72b-instruct & & 100 & 77 & 69 & 104.40 & 1.82 & 3.14 & 100 & 62 & 58 & 101.83 & 0.83 & 11.69 & 100 & 60 & 51 & 107.20 & 25.13 & 11.52\\
llama3.1-70b-instruct & & 93 & 61 & 52 & 115.13 & 6.22 & 12.55 & 100 & 54 & 54 & 100 & 0 & 3.64 & 91 & 57 & 53 & 109.21 & 80.49 & 10.50 \\
ChatGPT-4 & & 3 & 3 & 3 & 100 & 0 & 5.04 & 1 & 1 & 1 & 100 & 0 & 11.85 & 1 & 1 & 1 & 100 & 0 & 5.62 \\
\midrule
deepseek-V3 & AoP & 100 & 80 & 80 & 100 & 0 & 3.27 & 100 & 74 & 74 & 100 & 0 & 7.13 & 100 & 66 & 66 & 100 & 0 & 6.43 \\
qwen2.5-7b-instruct & A* & 100 & 57 & 55 & 101.16 & 1.19 & 3.03 & 100 & 36 & 34 & 102.97 & 8 & 7.85 & 99 & 25 & 24 & 104.49 & 36 & 7.13 \\
qwen2.5-32b-instruct & & 99 & 67 & 59 & 106.61 & 1.31 & 2.21 & 100 & 61 & 59 & 100.94 & 2.22 & 3.80 & 100 & 54 & 50 & 105.38 & 37.18 & 5.49 \\
qwen2.5-72b-instruct & & 99 & 77 & 73 & 101.57 & 0.36 & 1.89 & 100 & 62 & 62 & 100 & 0 & 3.00 & 100 & 61 & 60 & 100.23 & 1.04 & 4.89 \\
llama3.1-70b-instruct & & 100 & 82 & 81 & 100.30 & 0.05 & 2.62 & 100 & 74 & 73 & 100.63 & 1.94 & 4.18 & 100 & 70 & 67 & 102.06 & 29.77 & 10.02 \\
ChatGPT-4 & & 84 & 65 & 61 & 102.26 & 1.16 & 4.90 & 66 & 48 & 45 & 100.68 & 0.25 & 10.35 & 54 & 41 & 39 & 101.39 & 19.21 & 10.10 \\
\midrule
deepseek-V3 & AoP & 100 & 81 & 79 & 101.40 & 0.49 & 1.89 & 100 & 69 & 69 & 100 & 0 & 2.57 & 100 & 73 & 72 & 101.13 & 17.75 & 3.46 \\
qwen2.5-7b-instruct & Dijkstra & 100 & 57 & 55 & 1.0094 & 0.35 & 2.00 & 100 & 33 & 31 & 101.88 & 4.48 & 2.87 & 100 & 23 & 20 & 106.38 & 17.04 & 66.42\\
qwen2.5-32b-instruct & & 100 & 64 & 55 & 107.63 & 3.625 & 1.79 & 100 & 0.51 & 0.45 & 106.52 & 12.08 & 3.77 & 99 & 59 & 42 & 120.89 & 114.92 & 5.79 \\
qwen2.5-72b-instruct & & 100 & 78 & 75 & 101.05	& 0.56 & 2.09 & 100 &	62 & 61 & 100.54 & 0.26 & 3.98 &
1 & 66 & 66	& 1	& 0 & 5.89 \\
llama3.1-70b-instruct & & 96 & 67 & 64 & 103.44 & 1.25 & 1.50 & 95 & 52 & 51 & 101.55 & 4.92 & 2.85 &
98 & 63	& 61 & 103.81 & 14.29 & 4.34 \\
ChatGPT-4 & & 99 & 78 & 76 & 100.72 & 0.10 & 2.99 & 95 & 66	& 59 & 101.88 & 0.79 & 5.82 & 76 & 50 & 47 & 103.21 & 42.88 & 7.65 \\
\bottomrule
\end{tabular}
\caption{Evaluation metrics of different models with Algorithm of Planning (AoP) prompts}
\end{table*}
\begin{table*}[htbp]
\centering
\vspace*{-5.0em}
\label{tab:performance of AoT}
\scriptsize % 使用更小的字体
\setlength{\tabcolsep}{3pt} % 减少列间距
\begin{tabular}{@{}lc*{3}{cccccc}@{}}
\toprule
\multirow{2}{*}{Model} & \multirow{2}{*}{Prompt} & \multicolumn{6}{c}{10 size} & \multicolumn{6}{c}{20 size} & \multicolumn{6}{c}{30 size} \\
\cmidrule(lr){3-8} \cmidrule(lr){9-14} \cmidrule(lr){15-20}
 & & \multicolumn{1}{c}{CR} & \multicolumn{1}{c}{FR} & \multicolumn{1}{c}{OR} & \multicolumn{1}{c}{GM} & \multicolumn{1}{c}{MSE} & \multicolumn{1}{c}{RT(s)} & 
 \multicolumn{1}{c}{CR} & \multicolumn{1}{c}{FR} & \multicolumn{1}{c}{OR} & \multicolumn{1}{c}{GM} & \multicolumn{1}{c}{MSE} & \multicolumn{1}{c}{RT(s)} & 
 \multicolumn{1}{c}{CR} & \multicolumn{1}{c}{FR} & \multicolumn{1}{c}{OR} & \multicolumn{1}{c}{GM} & \multicolumn{1}{c}{MSE} & \multicolumn{1}{c}{RT(s)} \\
\midrule
deepseek-V3 & Easy & 100 & 84 & 82 & 100.54	& 0.10 & 4.79 & 100 & 76 & 74 & 101.61 &	5.26 & 6.22 & 100 & 71 & 70 & 101.56 & 3.61 & 8.29 \\
qwen2.5-7b-instruct & shot & 100 & 50 & 43 & 109.82 & 3.12 & 2.10 & 100 & 40 & 36 & 103.06 & 3.1 & 3.47 & 97 & 31 &	25 & 105.40 & 14.97 & 7.24 \\
qwen2.5-32b-instruct & & 100 & 71 & 58 & 109.22 &	3.83 & 2.04 & 99 & 54 & 49 & 106.75 & 22.44 &	3.67 & 98 & 59 & 51 & 105.12 & 19.53 & 5.00 \\
qwen2.5-72b-instruct & & 100 & 77 & 71 & 102.48 & 0.31 & 2.18 & 86 & 52 & 50 & 102.89 & 2.85 & 4.09 &
80 & 53 & 53 & 100 & 0 & 4.97 \\
llama3.1-70b-instruct & & 98 & 74 & 74 & 100 & 0 & 4.65 & 99 & 64 & 63 & 101.24 & 0.56 & 4.26 & 97	& 60 & 59 & 102.34 & 9.6 & 15.51 \\
ChatGPT-4 & & 85 & 69 &	69 & 100 & 0 & 10.89 & 74 & 49 & 46 & 102.22 & 10.29 & 13.74 & 62 & 44 & 44 & 100 & 0 & 17.47 \\
\midrule
deepseek-V3 & Complex & 100	& 82 & 79 & 100.82 & 0.15 & 2.06 & 100 & 67 & 67 & 100 &	0 &	2.91 & 100 & 64 & 62 & 100.96 & 3.81 & 4.06 \\
qwen2.5-7b-instruct & shot & 0 & 0 & 0 & NaN & NaN & NaN & 0 & 0 & 0 & NaN & NaN & NaN & 0 & 0 & 0 & NaN & NaN & NaN \\
qwen2.5-32b-instruct & & 95 & 39 & 33 & 110.15 & 2.77 & 2.65 & 93 & 49 & 41 & 110.00 & 19.27 & 5.78
& 98 & 39 &	0.3 & 112.20 & 72.51 & 7.09 \\
qwen2.5-72b-instruct & & 92	& 66 & 56 & 108.21 & 2.24 & 2.16 & 97 & 62 & 62 & 100 & 0 & 3.68 & 99 & 56 & 54 & 100.46 & 1.42 & 5.64 \\
llama3.1-70b-instruct & & 100 & 72 & 69 & 102.59 & 	0.33 & 1.57 & 100 & 62 & 59 & 103.62 & 15.61 & 1.96 & 100 & 56 & 55 & 100.64 & 2.57 & 11.69 \\
ChatGPT-4 & & 100 & 74 & 71 & 101.62 & 0.76 & 2.31 &
100	& 74 & 71 & 100.79 & 0.16 &	3.07 & 99 & 67 & 65	& 101.61 & 20.84 & 4.61 \\
\bottomrule
\end{tabular}
\caption{Evaluation metrics of different models with AoP with Example prompts}
\end{table*}

\clearpage

\begin{figure*}[ht!]
    \centering
    \includegraphics[width=\linewidth]{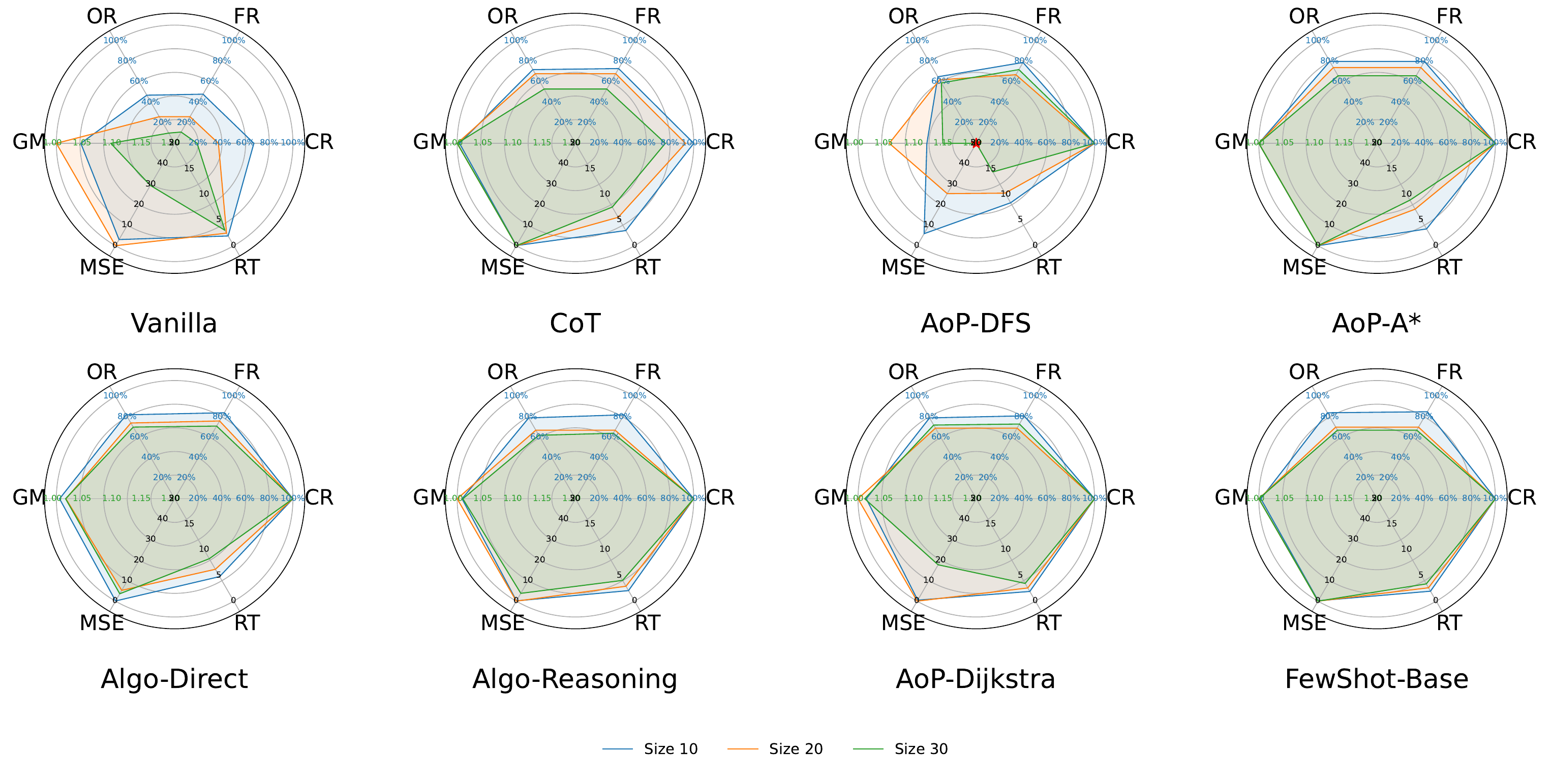}
    \caption{Deepseek-V3 performance}
    \label{fig:performance_metrics_cot}
\end{figure*}

\begin{figure*}[ht!]
    \centering
    \includegraphics[width=\linewidth]{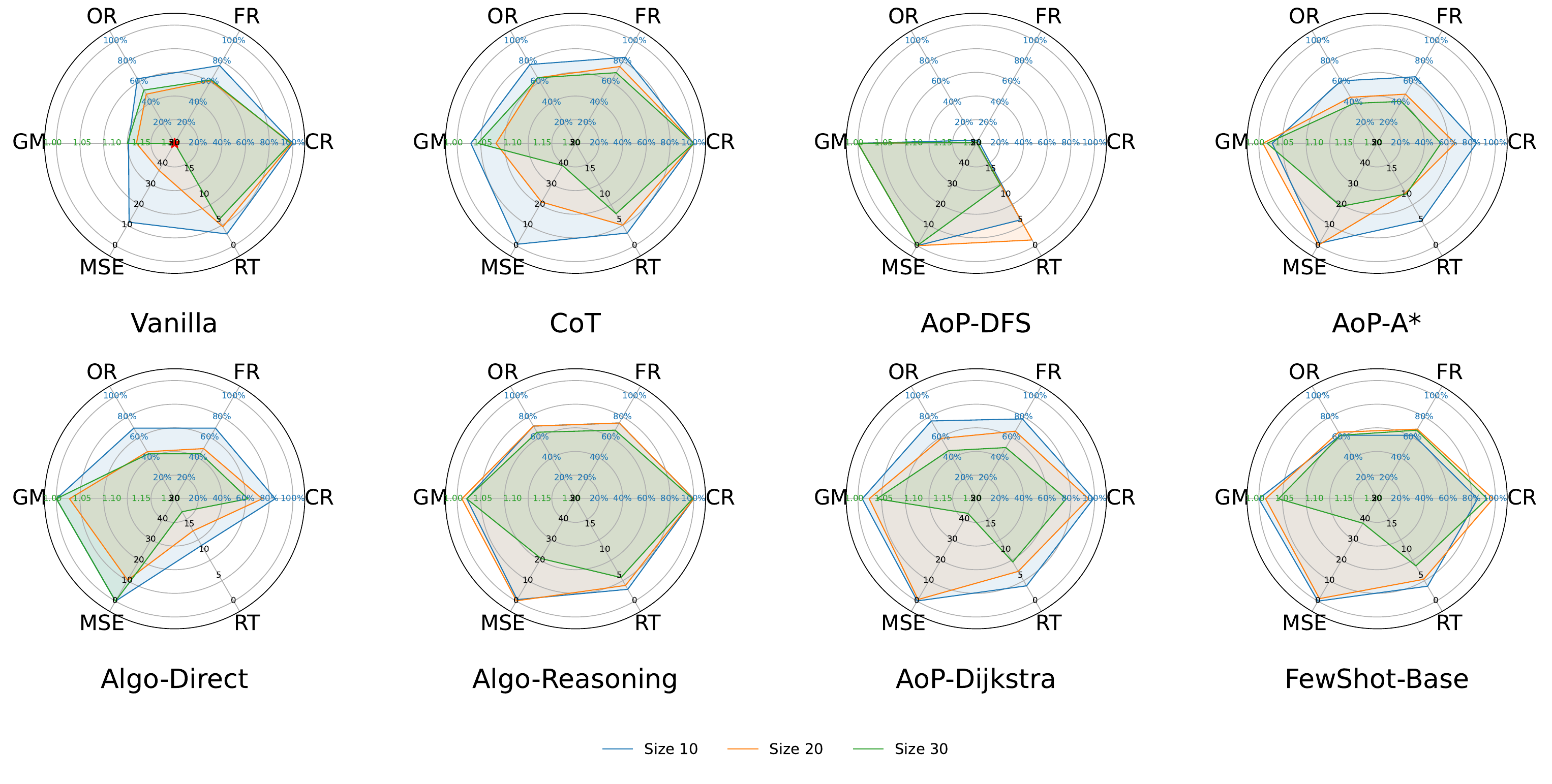}
    \caption{GPT4-Turbo performance}
    \label{fig:performance_metrics_cot}
\end{figure*}

\begin{figure*}[ht!]
    \centering
    \includegraphics[width=\linewidth]{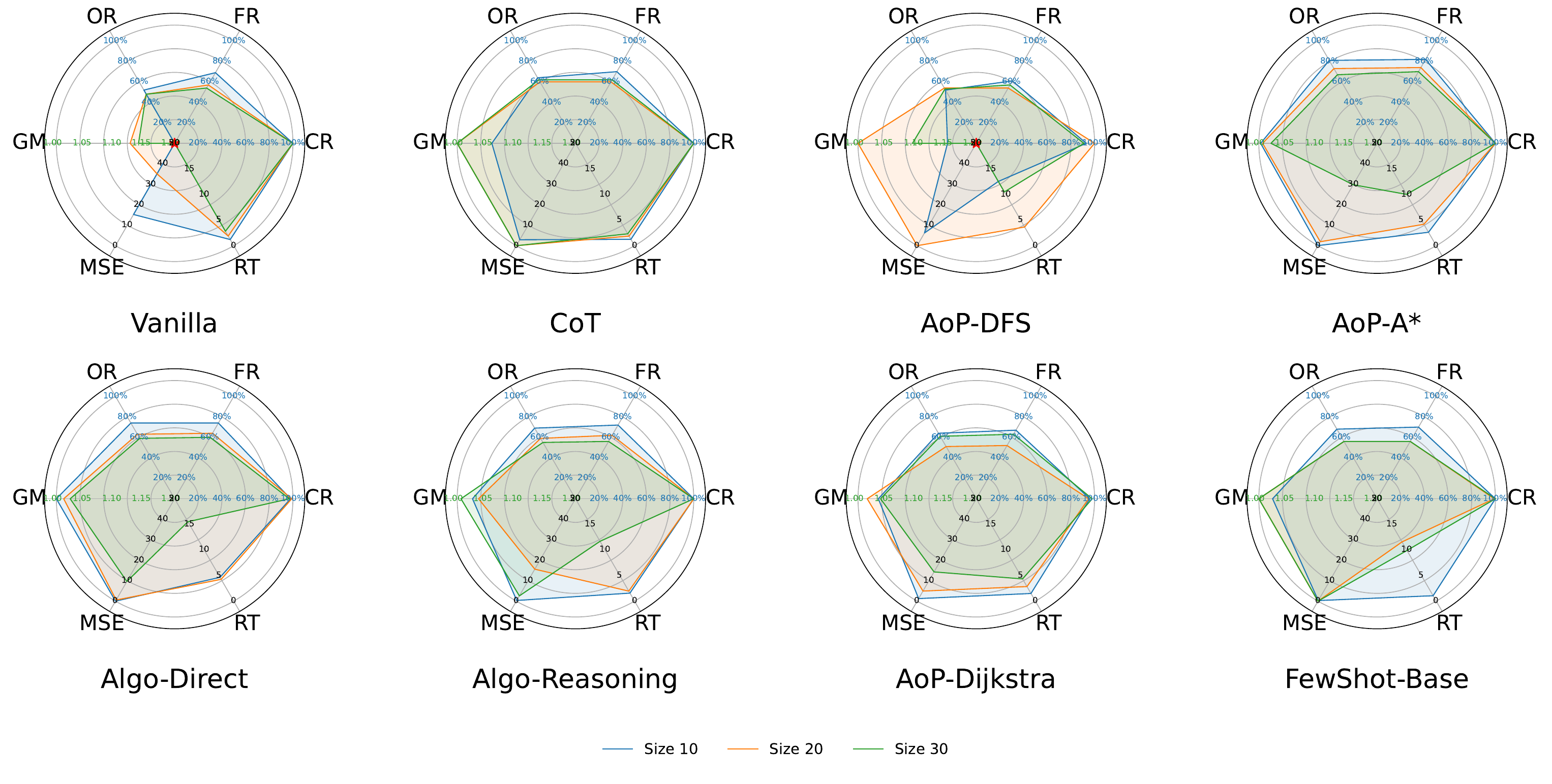}
    \caption{LLaMA3.1-70B performance}
    \label{fig:performance_metrics_cot}
\end{figure*}

\begin{figure*}[ht!]
    \centering
    \includegraphics[width=\linewidth]{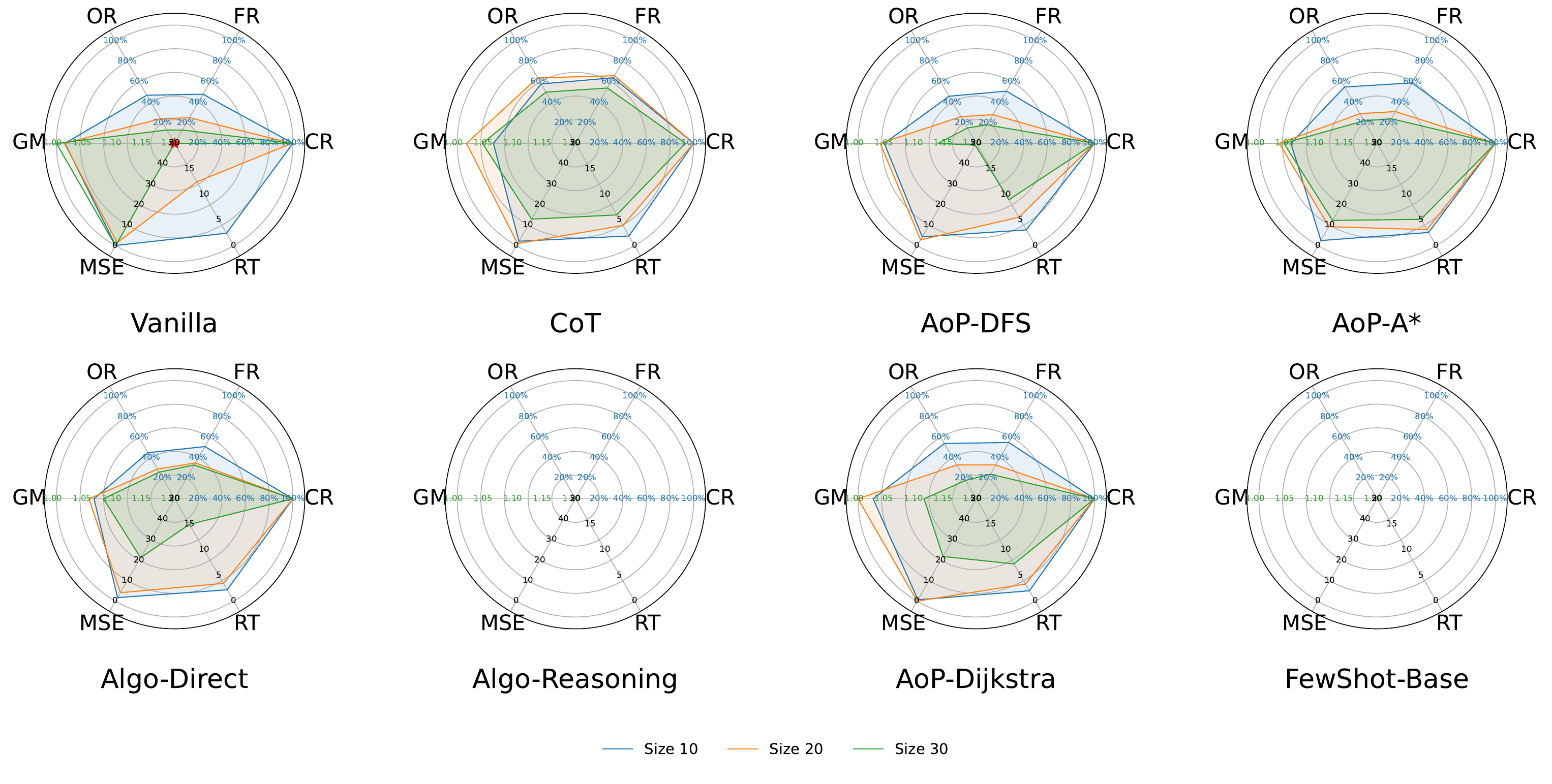}
    \caption{Qwen2.5-7B performance}
    \label{fig:performance_metrics_cot}
\end{figure*}

\begin{figure*}[ht!]
    \centering
    \includegraphics[width=\linewidth]{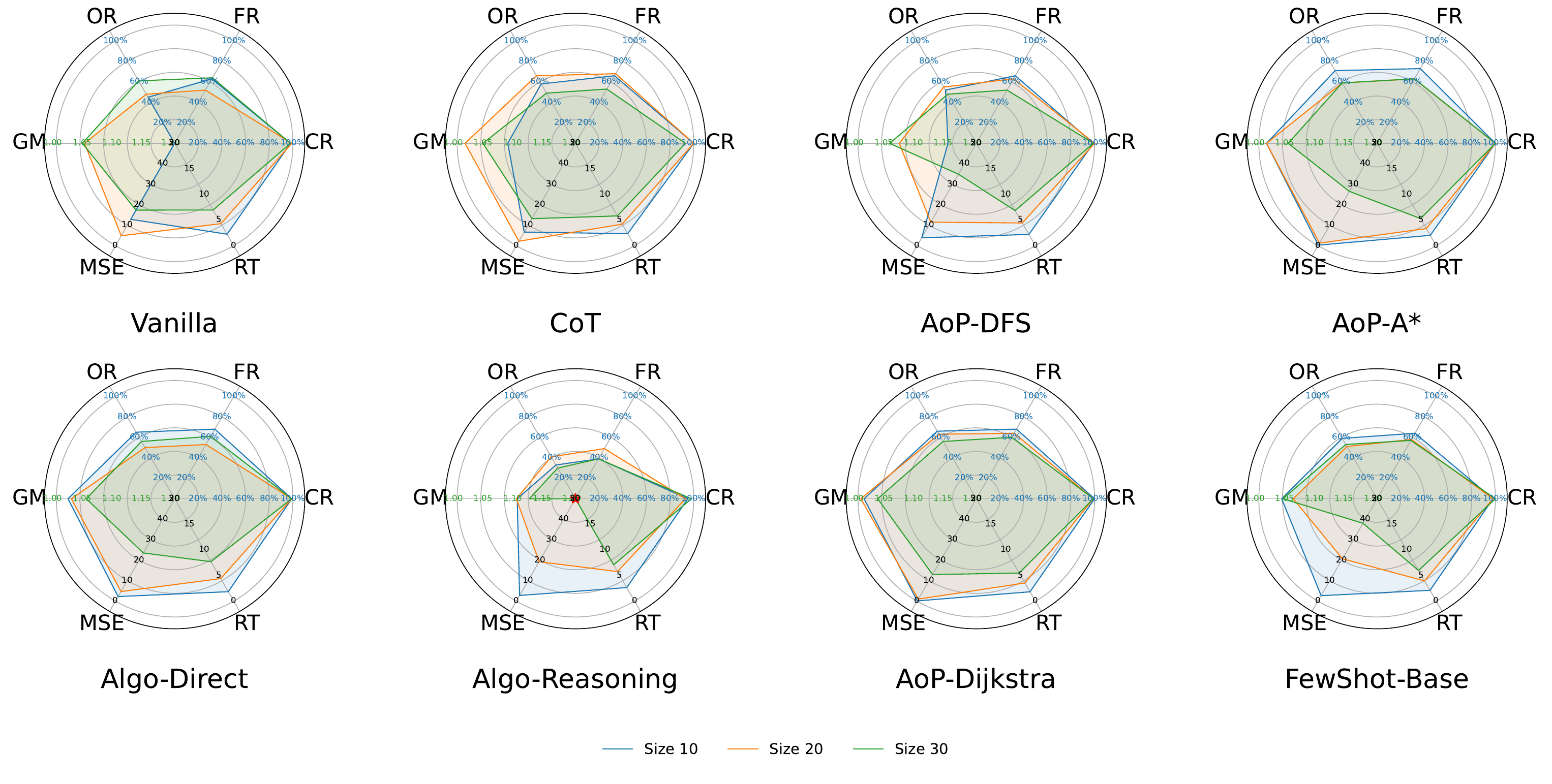}
    \caption{Qwen2.5-32B performance}
    \label{fig:performance_metrics_cot}
\end{figure*}

\begin{figure*}[ht!]
    \centering
    \includegraphics[width=\linewidth]{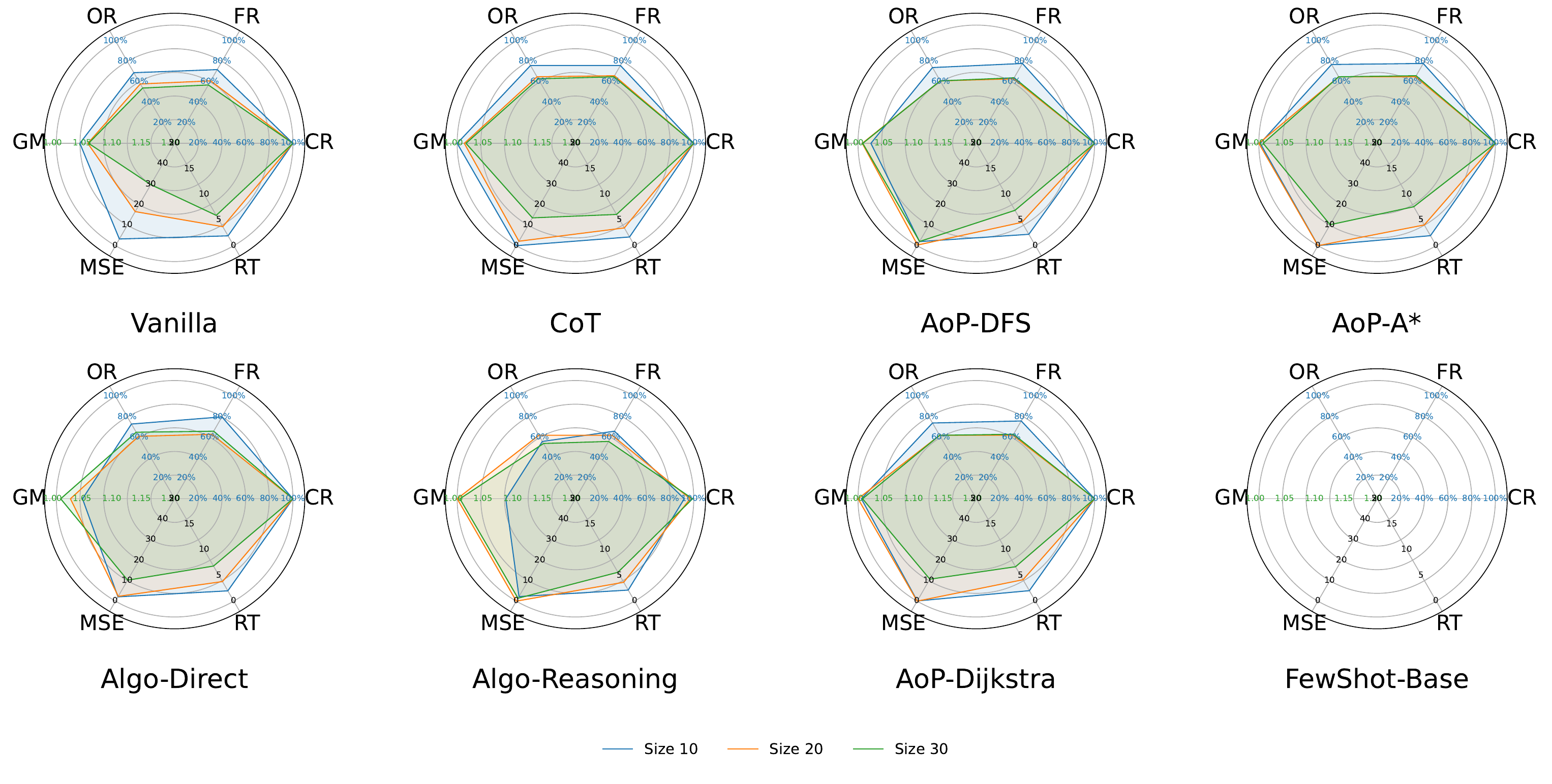}
    \caption{Qwen2.5-72B performance}
    \label{fig:performance_metrics_cot}
\end{figure*}
% \section{Example Appendix}
% \label{sec:appendix}

% This is an appendix.

\end{document}